# Deep Reinforcement Learning and Transportation Research: A Comprehensive Review

Nahid Parvez Farazi, Tanvir Ahamed, Limon Barua, Bo Zou[*]

Department of Civil, Material, and Environmental Engineering

University of Illinois at Chicago

**Abstract:** Deep reinforcement learning (DRL) is an emerging methodology that is transforming the way many complicated transportation decision-making problems are tackled. Researchers have been increasingly turning to this powerful learning-based methodology to solve challenging problems across transportation fields. While many promising applications have been reported in the literature, there remains a lack of comprehensive synthesis of the many DRL algorithms and their uses and adaptations. The objective of this paper is to fill this gap by conducting a comprehensive, synthesized review of DRL applications in transportation. We start by offering an overview of the DRL mathematical background, popular and promising DRL algorithms, and some highly effective DRL extensions. Building on this overview, a systematic investigation of about 150 DRL studies that have appeared in the transportation literature, divided into seven different categories, is performed. Building on this review, we continue to examine the applicability, strengths, shortcomings, and common and application-specific issues of DRL techniques with regard to their applications in transportation. In the end, we recommend directions for future research and present available resources for actually implementing DRL.

[*] Corresponding Author. Email: bzou@uic.edu.

# 1 Introduction

An important driving force of today's transportation research and development is the ideal of achieving fully connected and autonomous mobility. Towards this ideal, transportation operations are expected to witness a trend of reduced human involvement and increasingly lean towards artificial intelligence (AI). Daily transportation operations involve a wide range of complicated decision-making tasks. Deep reinforcement learning (DRL), which combines the power of deep learning (DL) in tackling large, complicated problems with reinforcement learning (RL) which provides a generic and flexible framework for sequential decision-making, brings a novel and powerful AI based tool that is gaining growing popularity in autonomous decision-making and operation control. In fact, with recent breakthroughs in DRL research, especially that DRL surpassed human level control in playing Atari games (Minh et al., 2015) and mastered the game of go by Silver et al. (2016, 2017), DRL has demonstrated its tremendous capability in solving an array of complicated decision-making tasks. The increasing availability of computational power serves as an additional catalyst that has broadened the horizon of DRL applications even further. As a result, like many other fields DRL is attracting a soaring attention from researchers in the transportation fields, where DRL algorithms have been increasingly used and adapted over the past few years. Many of these applications have produced significant results that outperform known benchmarks.

One salient feature of DRL is that high quality results can be generated at an extremely fast pace after the DRL model is trained, which is very crucial and desired in a connected and autonomous environment. Because of this, DRL based approaches are very attractive for real-time decision-making and control strategy designs such as in the context of autonomous driving (AD). Besides AD, DRL has been applied to fully adaptive traffic signal control (TSC), dynamic traffic management by road controls, and energy efficient driving. DRL also has a high potential for improving routing and vehicle dispatching, or more generally solving classical traveling salesman problems (TSP) and vehicle routing problems (VRP). Furthermore, DRL has witnessed successfully applications in rail and maritime transportation sectors. Fig. 1 shows the number of published papers in various transportation research domains from 2016 to 2020 (April). Given the mounting body of literature, what we find is missing is a comprehensive review and synthesis of the recent advances in employing and adapting DRL in transportation research.

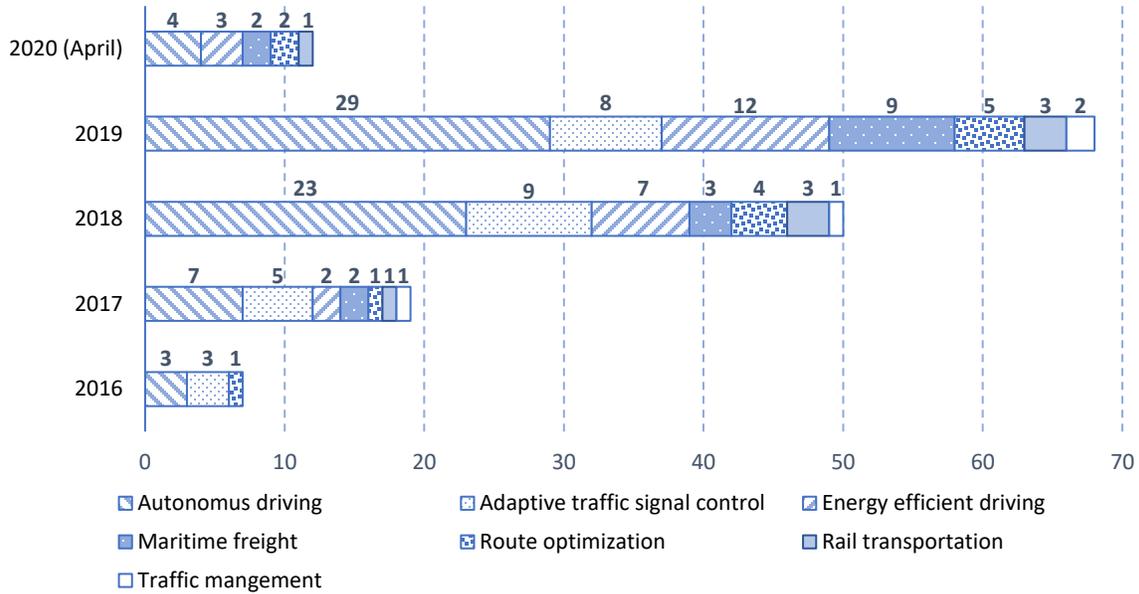

**Figure 1.** Number of papers using DRL in transportation from 2016 to 2020 (April).

This paper intends to fill this gap. We review and create a synthesis of DRL use in transportation research. To the best of our knowledge, this is the first effort to provide a comprehensive review of DRL applications throughout the transportation fields. The goal of this review is to expose a broader transportation audience to DRL in terms of its fundamentals, strengths, and weaknesses, synthesize what have been accomplished in existing research, identify what are the transportation problems that can be tackled using DRL, and explore directions for further use and adaptation of DRL in the transportation domain. To this end, in the rest of the paper we first offer an overview of the DRL methodological background. We start with the fundamentals of RL, and then move our discussion to how and where RL converges with DRL and what advantages DRL brings. Several DRL algorithms and some additional DRL extensions that are highly applicable and have potential to be applied in transportation are described. In the section describing DRL in transportation research, we divide our review of DRL applications into seven major segments: 1) autonomous driving; 2) adaptive traffic signal control; 3) energy efficient driving; 4) maritime freight transportation; 5) route optimization; 6) rail transportation; 7) traffic management system. Building on this detailed review, we conduct a synthesized discussion on the applicability of DRL based techniques in transportation research. Some major strengths and shortcomings of DRL as a method are assessed. We further examine common and application-specific issues of DRL techniques with regard to their transportation applications, and possible avenues for future research. The paper also offers a presentation of available resources for implementing DRL. Concluding remarks are given in the end.

# 2 Methodological background

This section presents the theoretical background of DRL. We first offer a brief description of basic reinforcement learning (RL), based on which we describe how RL is enhanced by integrating DL which gives birth to "Deep Reinforcement Learning". Both popular DRL algorithms that have been used in transportation research and some extensions of these algorithms are covered.

## 2.1 Reinforcement learning

RL is one of the three categories of machine learning (the other two are supervised learning and unsupervised learning) (Sutton and Barto, 2018). The focus of RL is to train an agent such that the agent can optimize its behavior by accumulating and learning from its experiences of interacting with the environment. The optimality is measured as maximizing the total reward by taking consecutive actions. More specifically, RL is a sequential decision process with the agent being the decision maker. At each decision point, the agent has information about the current state of the environment and selects an action that deems the most appropriate based on his experiences at that point. The action taken transitions the environment to a new state. Meanwhile, the agent gets some reward, i.e., reinforcement, as a signal of how good or bad the action taken is.

To formulate the sequential decision process, RL employs Markov Decision Processes (MDP) as the mathematical foundation to keep track of the progression of the decision process. To do so, the following notations are introduced. Set $S$ includes all possible states of the environment. Set $A$ contains all possible actions that the agent can take. Set $R$ includes all possible rewards as a result of the agent taking an action at a given state. To illustrate, the environment is in state $s_t$ at time step $t$. At this state, the agent received reward $r_t$ as a result of some immediate past action. Now the agent takes an action $a_t$. The action transitions the environment to a new state $s_{t+1}$ at the next time step $t+1$. Meanwhile, the agent receives a reward $r_{t+1}$ as a result of the action taken. The reward is a function of state-action pair: $\mathcal{R}(s_t, a_t) \to r_{t+1}$. This agent-environment interaction is further shown in Fig. 2.

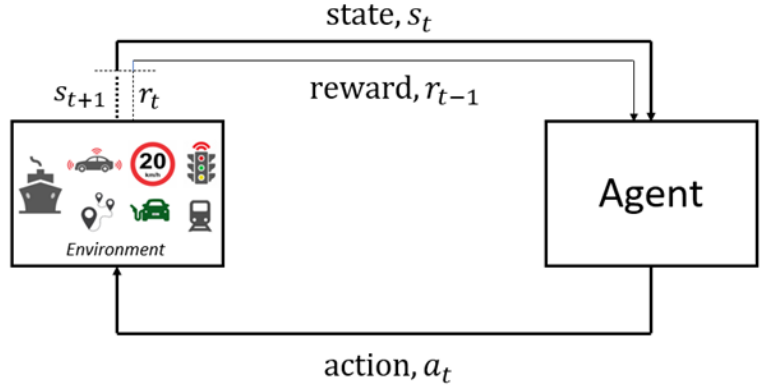

**Figure 2.** Illustration of agent-environment interaction.

Since the actions are taken sequentially, the objective of the agent is to maximize the cumulative reward, which is the expected return over the entire time period. At a time step $t$, the expected return $\bar{R}_t$ is the sum of rewards from the next time step till the last time step $T$:

$$\bar{R}_t = r_{t+1} + r_{t+2} + r_{t+3} + \cdots + r_T \tag{1}$$

If we consider that the reward is received over a long period, then a discount factor $\gamma$ should be incorporated to reflect discounting. In the case that a task does not have a terminal time, the expected return is:

$$\bar{R}_t = r_{t+1} + \gamma r_{t+2} + \gamma^2 r_{t+3} + \cdots = \sum_{k=0}^{\infty} \gamma^k r_{t+k+1} \tag{2}$$

*Exploration vs exploitation:* While taking an action, the agent needs to decide on two fundamental dilemmas – whether to take the best action based on the current information (exploit) or gather more information (explore). The most common approach to achieve this tradeoff between exploration and exploitation is $\varepsilon$ - greedy strategy. According to this strategy, the agent will take a random action with a probability $\varepsilon$. At the beginning of the training, $\varepsilon$ is set to 1 to ensure that the agent is exploring the environment. With time, $\varepsilon$ is reduced using a decay rate (decay rate is a hyperparameter) to make a tradeoff between exploration and exploitation. Such an exploration strategy is incorporated with one of the most popular DRL algorithms - deep Q-Network (DQN) (Minh et al., 2015).

Depending on the environment transition behavior, the RL algorithms can be classified into two classes: model-based and model-free. In model-based algorithms, given a state and action pair, a model (transition function) is used to predict the transitioned future state and resultant reward. Such algorithms decide on a course of action by considering anticipated future situations before they occur. In model-free algorithms, the agent does not need any model or transition function. Instead, this approach explicitly relies on a trial and error process (Sutton and Barto, 2018). The model-free algorithms are relatively simpler, inexpensive, and widely applied in transportation research. Thus, our methodological review of RL and DRL advancement is mainly focused on model-free approaches. Model-free algorithms can be further classified into two main classes: value-based algorithms, policy-based algorithms. There exist another hybrid class: actor-critic algorithms that combine both value-based and policy-based algorithms. Below we provide a brief overview of these three classes.

### 2.1.1. Value-based RL algorithms

Value-based RL algorithms involve estimating the value (i.e., expected return) of a given state at a given time. This is referred to as estimating the value function. A value function tells the agent how good it is for the agent to: 1) be in a state at a given time, or 2) take an action from a state at a given time. Accordingly, two types of value function exist: state-value function and action-value function. The state-value function $V_\pi(s)$, formally defined by Eq. (3), gives the expected return when the environment starts in state $s$ and follows policy $\pi = \pi(a|s)$ which is a mapping from states to probabilities of selecting each possible action.

$$V_\pi(s) = E_\pi(\bar{R}_t|s_t = s) = E_\pi\left(\sum_{k=0}^{\infty} \gamma^k r_{t+k+1} |s_t = s\right) \quad (3)$$

On the other hand, the action-value function $Q_\pi(s, a)$ is the expected return starting from state $s$, taking action $a$, and thereafter following under a policy $\pi$:

$$Q_\pi(s, a) = E_\pi(\bar{R}_t |s_t = s, a_t = a) = E_\pi\left(\sum_{k=0}^{\infty} \gamma^k r_{t+k+1} |s_t = s, a_t = a\right) \quad (4)$$

One of the most popular and widely used value-based RL algorithms is "Q-learning" (Watkins and Dayan, 1992). It is a model-free, off-policy approach. Here, off-policy means that regardless of the policy that enables the agent to take an action $a$ at the current state $s$, the agent will update the Q-value of the current state $s$ using the next state's optimal Q-value "$Q_*(s', a')$" where action $a'$ is a greedy action. The point is that the action $a$ of current state and action $a'$ of the next state are not from the same policy. This is ensured by the "max" operator in right hand side of the Eq. (5). On the contrary, in an on-policy approach, the action selection at the next state $s'$ would follow the same policy that enables the agent to take the action $a$ at the current state. Q-learning algorithm enables the agent to choose an action $a \in A$ with the highest Q-value available from a state $s \in S$ based on a Q matrix for any given time step. The Q matrix is a mapping for a discrete state and action space. This Q matrix is updated at every time step by following a Bellman optimality equation as shown in Eq. (5):

$$Q_{new}(s,a) \leftarrow (1-\alpha)Q_{old}(s,a) + \alpha \left[ r + \gamma \max_{a' \in A} Q_*(s',a') \right] \quad (5)$$

where $r$ is the obtained reward. $a' \in A$ denotes the action that gives the highest Q-value from the following state $s' \in S$. $\alpha$ represents the learning rate (hyperparameter) which takes values between 0 and 1.

### 2.1.2 Policy-based RL approach

Unlike the value-based approach, the policy-based approach does not require estimating the value of a certain state or state-action pair, rather it searches for an optimal policy $\pi^*$ directly. In policy-based approaches, typically a parameterized policy $\pi_\theta$ is chosen and this parameter $\theta$ of the policy $\pi_\theta$ is gradually updated to maximize the expected return as shown in the state-value function in Eq. (6). This parameter update can either be done by a gradient-free or gradient-based approach (Deisenroth et al., 2013). Gradient-based approaches are mostly used in existing DRL algorithms.

$$V_{\pi_\theta}(s) = E_{\pi_\theta}\left( \sum_{k=0}^{\infty} \gamma^k r_{t+k+1} | s_t = s \right) \quad (6)$$

The policy-based approaches are particularly suitable for very large or infinite action spaces. Let us consider $J(\theta)$ is a scalar performance measure that is to be maximized by updating the policy parameter $\theta$. Therefore, this update is done as:

$$\theta_{t+1} = \theta_t + \alpha \nabla \hat{J}(\theta_t) \quad (7)$$

Here, $\nabla \hat{J}(\theta_t)$ is a stochastic estimate of the gradient with respect to parameter $\theta$ (Sutton and Barto, 2018). REINFORCE (Williams, 1992) is a classical policy-based RL algorithm whose update involves only the taken action $a$ from a certain state $s$, and this update is done by:

$$\theta' \leftarrow \theta + \alpha \gamma^t \bar{R} \nabla \ln \pi_\theta(a|s,\theta) \quad (8)$$

Although the updating only requires the action $a_t$ taken at the current time step $t$, REINFORCE uses the complete expected return from the current time step. Therefore, this algorithm can be well defined only for the task with a terminal state with all the updates made after the episode is completed.

**2.1.3 Actor-critic approach**

The value-based algorithms cannot handle problems that involve continuous (real-valued), high-dimensional action spaces. In addition, since the agent learns to approximate the solution by the Bellman equation, the agent can resort to a near-optimal policy. On the other hand, in case of policy-based algorithms, gradient estimators may have large variances (Konda and Tsitsiklis, 2000). Moreover, with the change in policy, the new gradient is estimated irrespective of previous polices. Therefore, the agent is not learning with respect to the accumulation of previous information. To cope with this limitation, existing literature suggests embracing the actor-critic approach that combines both policy-based algorithm and value-based algorithm (Konda and Tsitsiklis, 2000; Grondman et al., 2012). In the actor-critic approach, the agent is trained using two estimators as shown in Fig. 3. One is known as the critic function, which approximates and updates the value function; the second one is known as the actor function, which controls the agent's behavior based on policy. Based on the value function derived from the critic function, the actor function's policy parameter is updated in the direction of performance improvement. While the actor function controls the agent's behavior based on policy, the critic function evaluates the selected action based on the value function. A recent actor-critic based DRL algorithm, termed as deep deterministic policy gradient (DDPG), is discussed in section 2.2.1.4.

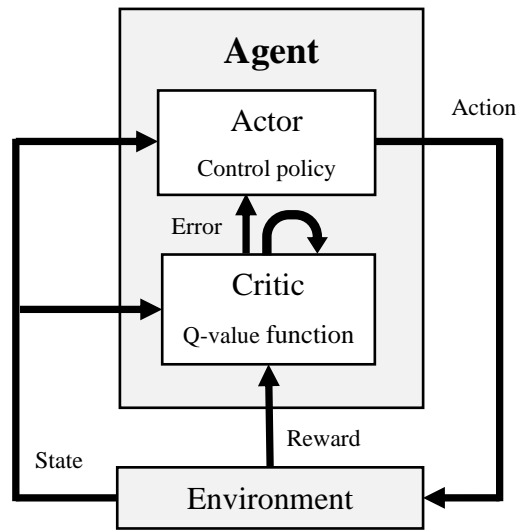

**Figure 3.** Conceptual framework for the actor-critic algorithm.

## 2.2 Deep reinforcement learning

In this section, we shift our discussion from RL to DRL. In principle, if the agent knows the optimal Q-value $Q_*(s, a)$ for every state-action pair, then the objective of RL is achieved. From every state, the agent will always find the highest Q-value from the Q matrix and choose the corresponding action. Such a table of Q-value can only be developed by recursively solving Eq. (5) with a small state and action space. However, for a large state and action space, such tabular format becomes computationally inefficient and even infeasible. Each additional feature added in the state space leads to an exponential growth in the number of Q-values that need to be stored in the table (Sutton and Barto, 2018). The real-world problems are most likely to be associated with a very large state space with a high dimension. To mitigate this curse of dimensionality, DL can be integrated with RL as a function approximator. A parameterized deep neural network (DNN) can approximate the optimal Q-values instead of computing the Q-values directly using Eq. (5). Different types of artificial neural network such as convolutional neural network (CNN), recurrent neural network (RNN) can be integrated with RL to deal with a very large state and action space (LeCun et al., 2015).

### 2.2.1 DRL algorithms

In this section, we describe some of the popular DRL algorithms that are frequently applied in transportation research. DQN and DDPG are the two most popular algorithms among transportation researchers. Two variants of DQN - double DQN and dueling DQN are also frequently applied. Furthermore, we discuss another recent algorithm PPO that is increasingly applied in transportation. Besides these five, there exist lots of other algorithms in the literature, we mention some of them in section 5.

#### 2.2.1.1 Deep Q-network (DQN)

Proposed by Minh et al. (2015), DQN uses a DNN (e.g., CNN) as the function approximator to approximate Q-value associated with a state-action pair. During training, at each time step the DQN algorithm follows the $\varepsilon$-greedy strategy (described in section 2.1) to make a tradeoff between exploration and exploitation while deciding on an action. One salient feature of training with DQN is experience replay, which involves a replay memory ***M*** that stores the agent's experiences during training. The replay memory can store experiences up to a certain limit which is to be initialized at the beginning of the training. An experience is associated with the agent taking an action at a given state and time step, observing the state transition, and getting a reward. Therefore, the experience is denoted as $e_t = (s_t, a_t, r_t, s_{t+1})$. Once the memory of $M$ is full, the oldest experience in $M$ is removed to create space for new experience.

The DNN is trained from a minibatch $U(M)$ of a randomly selected sample (state-action reward combinations) from $M$. The employment of experience replay with minibatch sampling brings several advantages. First, the learning from random samples results in less correlation compared to learning directly from consecutive samples, which increases the learning efficiency. Second, experience replay gives greater data efficiency by allowing each experience to be used in many weight updates. Third, by averaging the behavior distribution over many previous states, experience replay contributes to smoothing out learning and avoiding oscillation or divergence in the parameters (Mnih et al., 2015).

After assigning a random weight $\theta$ to the DNN, for each experience the input (state $s$) is allowed to propagate through the DNN. Now, the output $Q(s, a; \theta)$ is compared with the target optimal Q-value $Q_*(s_t, a_t)$ to compute the loss. Ideally, the expected value of this target Q-value should satisfy the Bellman optimality equation which would be equal to $r + \gamma \max_{a' \in A} Q_*(s', a')$. Therefore, the loss function can be written as:

$$\mathcal{L}(\theta) = E_{s,a,r,s' \sim U(M)} \left[ \left( r + \gamma \max_{a' \in A} Q_*(s', a') - Q(s, a: \theta) \right)^2 \right] \quad (9)$$

In Eq. (9), the right-hand side has an unknown part $Q_*(s', a')$ which denotes the optimal Q-value of the next iteration. One way to approximate this unknown quantity is to do a second forward pass with the DNN, i.e., for state $s'$ in each experience $(s, a, r, s')$ along with the same weight parameters $\theta$ of the DNN, predict state-action values $Q(s', a': \theta), \forall a' \in A$ using the DNN. Assuming $Q(s', a'; \theta) \approx Q_*(s', a')$, the loss value can be calculated using the following equation:

$$\mathcal{L}(\theta) = E_{s,a,r,s' \sim U(M)} \left[ \left( r + \gamma \max_{a' \in A} Q_*(s', a': \theta) - Q(s, a: \theta) \right)^2 \right] \quad (10)$$

After performing these two forward passes, the gradient of loss from Eq. (10) is used to update the weight $\theta$ of the DNN. However, there is a major drawback in these two forward pass procedure. Since the second forward pass is done in the same network with the same network parameter $\theta$, both Q-values and target Q-values will update in the same direction with the update of the weight parameter $\theta$. As a result, the correlation between the Q-values and the target Q-values can be high, which may cause oscillation or divergence of the policy during training.

To tackle this issue, Minh et al. (2015) propose a novel technique of creating a parallel network called the target network. The target network is structurally cloned as the original DNN. The target network parameter $\theta'$ is initialized as the same parameter as the original DNN $\theta$ at the beginning of the training. Unlike DNN parameter $\theta$, the target network parameter $\theta'$ is kept frozen for a certain number of iterations before updating. The number of iterations ($\delta$) the weights are kept frozen is a hyperparameter of the target network. After every $\delta$ iterations, the target network parameter $\theta'$ is updated to whatever the DNN parameter $\theta$ at that time. Finally, the loss functions can be written as:

$$\mathcal{L}(\theta) = E_{s,a,r,s' \sim U(M)} \left[ \left( r + \gamma \max_{a' \in A} Q_*(s', a': \theta') - Q(s, a: \theta) \right)^2 \right] \quad (11)$$

In the above Eq. (11), the first term of the right-hand side is known as the target Q-value, $y^{DQN} = r + \gamma \max_{a' \in A} Q_*(s', a': \theta')$. Now, the gradient of this loss is used to adjust the DNN parameter $\theta$ by backpropagation and gradient descent.

While sampling from the replay memory, the experience replay technique randomly creates a uniform minibatch of experiences to update the DNN parameter. This sampling technique selects the experiences at the same frequency as they originally occur, regardless of their significance. However, sometimes it is necessary to sample some specific experiences more frequently so that the learning process can become more efficient. For instance, in autonomous driving, one would expect to have fewer collision experiences. However, this smaller number of collision experiences will result in less exposure to collision scenarios during the training phase. This necessitates some prioritization of experience sampling. Schaul et al. (2015) propose a "prioritized experience replay" technique to tackle this issue. This technique prioritizes the experiences based on the magnitude of their temporal difference error and can double the learning speed.

### 2.2.1.2 Double DQN

Double DQN (van Hasselt et al., 2016) is a modified version of the DQN algorithm. This algorithm generalizes the tabular double Q-learning algorithm (van Hasselt et al., 2010) with a nonlinear function approximator such as the deep neural network. In the DQN, the target Q-value is calculated using the following equation:

$$y^{DQN} = r + \gamma \max_{a' \in A} Q(s', a'; \theta') \quad (12)$$

In this target Q-value, both the greedy action selection and action evaluation is done using the same network with parameter $\theta'$. This leads to a selection of overestimated policies that in turn results in overoptimistic value estimates. The idea of the double DQN algorithm is to decouple this action evaluation from the action selection. This is done by decomposing the max operator into action selection from the action evaluation. Without introducing any additional network, the double DQN algorithm utilizes the target network (integrated with DQN) as a second value function approximator. The target Q-value in double DQN algorithm is estimated as:

$$y^{DoubleDQN} = r + \gamma Q\left(s', \operatorname*{argmax}_{a' \in A} Q(s', a'; \theta); \theta'\right) \qquad (13)$$

According to Eq. (13), the evaluation of the greedy policy is done by the Q-network (parameter $\theta$) within the argmax operator and its value is estimated using the target network (parameter $\theta'$). The numerical experimentation in van Hasselt et al. (2016) shows that double DQN can find better policies and provide more stable learning than DQN.

### 2.2.1.3 Dueling DQN

Dueling DQN is a modified version of the DQN algorithm that follows the dueling network architecture (Wang et al., 2016). The convolutional layers in dueling architecture are similar to DQN except it has two sequences of fully connected convolution layers instead of one single sequence. These two sequences estimate the state value function and the advantage function separately. In RL, the advantage function estimates the degree to which the expected total discounted return is increased by performing an action rather than the action currently considered best (Harmon et al., 1995). In other words, advantage function $A^\pi(s, a)$ for an action $a \sim \pi(s)$ from a state $s$ is the difference between the Q-value associated with this state-action pair $Q^\pi(s, a)$ and the state value function $V^\pi(s)$.

Combining the estimates of the state value function and advantage function from two separate sequences, a single Q function is estimated as output as follows:

$$Q(s, a; \theta, \alpha, \beta) = V(s; \theta, \alpha) + \left(A(s, a; \theta, \alpha) - \frac{1}{|A|} \sum_{a'} A(s, a'; \theta, \alpha)\right) \qquad (14)$$

where $\alpha$ and $\beta$ are the parameters of the two fully connected layers. The key idea behind such architecture is to avoid unnecessary estimation of the value of every action choice from a certain state. Sometimes, just knowing the value of the state suffices for the agent to identify the best action to take without knowing the individual value for every action choice. Dueling DQN can be combined with double DQN. Double dueling DQN with prioritized experience replay is the most updated variant of the DQN algorithm.

### 2.2.1.4 Deep deterministic policy gradient (DDPG)

At this point, we refer the reader to a recently proposed actor-critic based deep deterministic policy gradient (DDPG) algorithm (Lillicrap et al., 2015) that has garnered popularity since it can operate over continuous action spaces. Based on the deterministic policy gradient algorithm (Silver et al., 2014), DDPG employs a parameterized actor function (stores current policy) with a parameterized critic function that approximates (using Bellman equation as in Q-learning) and updates the value function using samples, thereby tackling the large variance in the policy gradients of actor-only methods. The authors propose several modifications to the deterministic policy gradient that are inspired by the success of the DQN. Most notably, a recent advance in DL, termed as batch normalization (Ioffe and Szegedy, 2015) is introduced and reported to learn competitive policies. Since the state space can incorporate various information in terms of units and their ranges can fluctuate abruptly during training, it will be difficult for the network to learn, especially from low-dimensional state spaces. The proposed technique normalizes each element of the state spaces across the samples in a minibatch and maintains a unit mean and variance, thereby minimizing the covariate shift (i.e., the change in the distribution of the input variables) to accelerate the learning during training (Li, 2017).

The algorithm enables the agent to interact with the environment and employ a gradient descent to improve the policy using a minibatch collected from the replay memory. Using the sampled policy gradient, we can update the actor policy using Eq. (15).

$$\nabla_{\theta^\mu} J \approx \frac{1}{N} \sum_{i=1}^{N} \nabla_a Q(s, a|\theta^Q)|_{s=s_i, a=\mu(s_i)} \nabla_{\theta^\mu} \mu(s|\theta^\mu)|_{s_i} \tag{15}$$

where $N$ represents the size of the minibatch and $\mu(s|\theta^\mu)$ denotes the selected action based on the current policy. At the beginning of the training, the parameters of actor and critic network are initialized as $\theta^\mu$ and $\theta^Q$, respectively. Additionally, the target networks are denoted as $\mu'$ and $Q'$ with their weights as $\theta^{\mu'}$ and $\theta^{Q'}$, respectively. In contrast to the actor network, the critic network is updated using gradient descent on the loss function $\mathcal{L}(\theta^Q)$ as given below.

$$\mathcal{L}(\theta^Q) = \frac{1}{N} \sum_{i=1}^{N} (y_i - Q(s_i, a_i|\theta^Q))^2 \tag{16}$$

$$y_i = r_i + \gamma Q'(s_i', \mu(s_i'|\theta^{\mu'})|\theta^{Q'}) \tag{17}$$

Finally, the algorithm suggests employing two target networks cloning the actor and critic networks. However, instead of directly copying the polices (as done in DQN), this algorithm uses a soft target update using the Eq. (18) and Eq. (19), where $\tau \ll 1$. The soft target update strategy offers a more stable learning framework. The risk of unstable learning arises from the fact that the network being updated is also used in calculating the target value. The proposed soft target update is performed at a regular interval, using duplicate actor and critic networks by slowly maintaining the learned networks (target values).

$$\theta^{Q'} \leftarrow \tau\theta^Q + (1-\tau)\theta^{Q'} \tag{18}$$

$$\theta^{\mu'} \leftarrow \tau\theta^\mu + (1-\tau)\theta^{\mu'} \tag{19}$$

#### 2.2.1.5 Proximal policy optimization (PPO)

The success of policy gradient methods largely depends on the choice of the step size. If a step size is too small, then the progress is very slow and can take up to millions of steps to learn a very simple task. On the other hand, a large step size can often lead to a destructively large policy update that may catastrophically drop the performance. Proximal policy optimization (Schulman et al., 2017) is a new type of policy gradient algorithmic technique developed by OpenAI that shows promising results in continuous action space. This new method uses a stochastic gradient ascent and a novel objective function that enables multiple epochs of minibatch gradient updates. PPO method attempts to address the issue of step size in the policy gradient methods by combining the trust region policy optimization (TRPO) (Schulman et al., 2015) method with a clipped objective function. If we denote $\rho(\theta) = \frac{\pi_\theta(a|s)}{\pi_{\theta_{old}}(a|s)}$ as the probability ratio where $\pi_\theta$ refers to a stochastic policy, TRPO uses the maximization objective function:

$$\mathcal{L}^{CFI}(\theta) = \hat{E}[\rho(\theta)\hat{A}] \tag{20}$$

where the expectation $\hat{E}$ denotes the empirical average over a finite batch of samples and $\hat{A}$ denotes an estimator of the advantage function at the current time step (Schulman et al., 2017). The term $\mathcal{L}^{CFI}(\theta)$ denotes the conservative policy iteration (Kakade and Langford, 2002). Without any step size constraint, the objective function will result in a very large policy update (with the increase of $\pi_\theta(a|s)$, the objective function of Eq. (20) will also increase, this results in large policy update). The PPO methods address this issue by modifying this objective function, the so-called clipped surrogate objective function associated with PPO can be written as:

$$\mathcal{L}^{CLIP}(\theta) = \hat{E}\left[\min\left(\rho(\theta)\hat{A}, \text{clip}(\rho(\theta), 1-\epsilon, 1+\epsilon)\hat{A}\right)\right] \tag{21}$$

where $\epsilon$ is a hyperparameter of PPO algorithm that takes value from 0 to 1. The first part of the right-hand side is the unclipped portion (same as TRPO, Eq. (20)) and the second part is the clipped portion which removes the incentive for moving $\rho(\theta)$ outside the bound $[1-\epsilon, 1+\epsilon]$. The min operator in the objective function creates a pessimistic bound which essentially ensures that the clipping is only considered when it makes the objective worse than TRPO (Schulman et al., 2017). Finally, if this PPO algorithm is incorporated with an actor-critic architecture that shares parameters with policy and value function, the loss function will take the following form:

$$\mathcal{L}^{CLIP+VF+S}(\theta) = \hat{E}[\mathcal{L}^{CLIP}(\theta) - c_1\mathcal{L}^{VF}(\theta) + c_2 S[\pi_\theta](s)] \tag{22}$$

where $c_1$, $c_2$ are coefficients, $\mathcal{L}^{VF}(\theta)$ is the squared-error loss of value function $\left(V_{\pi_\theta}(s) - V_{target}\right)^2$, $S$ refers to the entropy loss (entropy loss is meant for assisting in the exploration). The numerical experiment results illustrated in Schulman et al. (2017) show that the PPO algorithm can outperform other similar algorithms such as Advantage Actor-Critic (A2C) and Asynchronous Advantage Actor-Critic (A3C) in actor-critic architecture. This type of algorithm is usually easy to implement, achieves ease of sample complexity and hyperparameter tuning in the continuous control domain.

### 2.2.2 Some extensions in DRL
#### 2.2.2.1 Multi-agent systems

Incorporating multi-agent systems in DRL algorithms, it is possible to enable multiple agents to interact with the environment and learn simultaneously. The environment in a multi-agent system cannot be considered as stationary. Since a non-stationary environment cannot retain the Markov property that indicates that the future dynamics of the environment is not just dependent on the current state (Tan, 1993; Laurent et al., 2011; Nowé et al., 2012), multi-agent RL and DRL problems are usually formulated as a simple stochastic game (also known as Markov game) (Littman, 1994; Busoniu et al., 2010). The stochastic game associated with the multi-agent DRL is a tuple $(S, \boldsymbol{A_1}, \ldots, \boldsymbol{A_N}, P, R_1, \ldots, R_N)$, where $N$ is the number of agents in the multi-agent system; $S$ denotes the state space of the environment; $\boldsymbol{A_i}, i = 1, \ldots, N$ is the finite set of actions available to the agent $i$; $P$ is the state transition probability function; and $R_i, i = 1, \ldots, N$ are the reward functions for the agent $i$.

Now, this gives us a joint action set $\boldsymbol{A} = \boldsymbol{A_1} \times \boldsymbol{A_2} \times .. \times \boldsymbol{A_N}$ and thus, state transition probability function can be defined as $P: S \times \boldsymbol{A} \rightarrow S$.

At a certain time step $t$, the transition of state and rewards depend on the joint action taken by all the agents in the system together $\boldsymbol{a}_t = a_{1,t} \times a_{2,t} \times .. \times a_{N,t}$, where $\boldsymbol{a}_t \in \boldsymbol{A}$ and $a_{i,t} \in \boldsymbol{A}_i$. For a joint policy $\boldsymbol{\pi} = \pi_1 \times \pi_2 \times .. \times \pi_N$, the expected value of the state value function for agent $i$ at time step $t$ can be expressed as:

$$V_{i,t}^{\boldsymbol{\pi}}(s) = \mathrm{E}^{\boldsymbol{\pi}} \left( \sum_{k=0}^{\infty} \gamma^k r_{i,t+k+1} | s_t = s \right) \qquad (23)$$

The Q-function for each agent also depends on the joint action and joint policy:

$$Q_{i,t}^{\boldsymbol{\pi}}(s, \boldsymbol{a}) = \mathrm{E}^{\boldsymbol{\pi}} \left( \sum_{k=0}^{\infty} \gamma^k r_{i,t+k+1} | s_t = s, \boldsymbol{a}_t = \boldsymbol{a} \right) \qquad (24)$$

Depending on the interactions among the agents, the multi-agent system can be fully cooperative (or collaborative), fully competitive (or adversarial), and mixed (Buşoniu et al. 2010; Hernandez-Leal et al., 2019). In a fully cooperative stochastic game, agents' reward functions (or goals) are the same. If the agents operate with opposing goals, then the game is considered as fully competitive. The mixed games have elements of both types. Multi-agent systems can represent the whole problem in a decentralized manner. Identifying each contributing party with an individual agent and considering each of their decision separately will give a more comprehensive and distributed perspective of the whole system.

### 2.2.2.2 Hierarchical DRL

One of the major drawbacks of RL is the curse of dimensionality (Barto and Mahadevan, 2003). Although the DRL can deal with large state and action spaces, still challenge remains to do it in a sample efficient manner. By creating hierarchies of policies, hierarchical RL (HRL) increases both the efficiency of learning and overall performance. By dividing the policies into several sub-policies, the action space becomes smaller that aids in better exploration of the environment. It also creates opportunities for transfer learning, especially at the lower level. One of the early works of the HRL framework is Feudal RL (Dayan and Hinton, 1993) where a hierarchy of managers is demonstrated. In this framework, a level of managers can control some sub-managers, at the same time these managers are controlled by some super-managers. Each manager assigns a goal for each of its sub-managers.

Another early work of HRL is the options framework by Sutton et al. (1999). This approach allows higher level policies to focus on goals and the lower level sub-policies are mainly focus on learning of controls. With the advent of DRL research, several new hierarchical DRL algorithms are proposed that are inspired by the early works of Dayan and Hinton (1993) and Sutton et al. (1999). A deep RNN based framework is proposed by Vezhnevets et al. (2016) to learn the macro (high) level policies in DRL. Vezhnevets et al. (2017) propose FeUdal Networks for HRL that takes long short-term memory (LSTM) network as the baseline on top of a representation learned by a CNN. Kulkarni et al. (2016) integrate DQN with HRL to propose the h-DQN framework.

#### 2.2.2.3 Asynchronous DRL

Asynchronous method is a special type of computing process where multiple parallel processes happen independently. One of the problems of using deep neural networks as the function approximator in DRL is the correlation between the updates that come from the sequential process. Minh et al. (2015) propose to utilize experience replay in the DQN algorithm to mitigate the effect of this correlation. In this method, parameter updates are achieved from a randomly sampled batch of experiences from the replay memory instead of using the experience immediately after it generates. However, the process of replay memory takes a toll on memory and computation power, and this off-policy learning uses data that is generated using an older policy (Minh et al., 2016).

Parallel computing process has the potential to increase the learning efficiency due to the increase of recent computational power, especially by multiple core CPU and GPU. Proposed by Minh et al. (2016), asynchronous advantage actor-critic (A3C) is an asynchronous DRL algorithm in actor-critic architecture. In this algorithm, several actor-critic agents learn parallelly and independently. Every agent acts on a different part of the environment with a different set of parameters, thus the changes in the parameters made by multiple agents in parallel are likely to be less correlated. The updates from every agent are received by a global network and combined asynchronously to achieve a global policy. Although multiple agents (oftentimes called workers to avoid confusion) works parallelly in normal A3C framework, they are not in the domain of multi-agent RL as the agents here are independent and do not interact. However, it is possible to allow interactions among several agents in the A3C framework; in that case, it becomes a multi-agent A3C algorithm. A synchronous and deterministic variant of A3C is known as advantage actor-critic (A2C) proposed by OpenAI gym. The asynchronous framework is suitable for deep neural network training for both value-based and policy-based methods, training can be achieved both on-policy and off-policy approach and can deal with both discrete and continuous action spaces. Minh et al. (2016)'s experimental results on A3C show that it can achieve faster training time compared to DQN.

#### 2.2.2.4 Imitation learning and inverse RL

Imitation learning is a process of learning from demonstrations which is also known as "apprenticeship learning". The idea is what if the agent has no idea what the reward is, how the agent can approximate its behavior to find the best policy without any reward to guide. In this case, with a set of expert demonstrations (typically defined by humans), the agent tries to learn the optimal policy imitating the expert's decisions. Usually, the expert demonstrations are provided in the form of some trajectories ($\tau = s_0, a_0, s_1, a_1, ....$), where these trajectories are built on some good policies. Depending on the loss function and the learning algorithm, there can be variations in the approach of imitation learning. One way to learn from the expert demonstration is to extract the reward signal which is known as Inverse RL (Ng and Russell, 2000). In inverse RL, the agent first learns a reward signal from the expert demonstrations, and then uses this reward function to find the optimal policy. Through inverse RL, sometimes it is difficult to extract a unique reward signal, multiple reward signals can result in the same reward value for a given policy. However, it is possible to find several candidate reward signals. With some strong assumptions, including thorough knowledge of the environment dynamics and completely solving the problem several times, a proper reward signal can be extracted to find the optimal policy (Sutton and Barto, 2018). Inverse RL can be applied in a DRL framework by incorporated deep neural networks as a function approximator. For more comprehensive readings on imitation learning and inverse RL in the DRL framework, we recommend readers the following literature (Stadie et al., 2017; Hester et al., 2018; Wulfmeier et al., 2015).

In many real-world problems, it is very difficult to derive a proper reward function. The successful completion of a task may depend on multiple factors that require some parameterization of the reward function. Weights of these parameters need to be defined manually which involves an intense tuning process. For instance, the success of lane change not only depends on whether the agent successfully completes the task but also depends on the marginal safety risk, smoothness, and comfortability of the transition. It is difficult to fix the weight for each of these factors in the reward function beforehand without knowing the marginal effect of any changes in the weight on the agent's behavior. Similarly, any task that requires to optimize multiple factors through a common reward signal can utilize inverse RL to ease the overall learning efficiency.

**2.2.2.5 Attention and pointer network**

Earlier advancements in the sequence-to-sequence learning (Sutskever et al., 2014) can be attributed to the enchanting prediction abilities of neural networks, which has motivated application in a broader array of combinatorial optimization problems, such as the travelling salesman problem (TSP). However, subsequent research work (Bahdanau et al., 2014) confirms that the agent is required to be trained all over again for variable-sized problems to decode the sequence of the locations to be visited. To mitigate this hindrance, TSP is investigated using a pointer network concept that is introduced as in Vinyals et al. (2015) owing to its superior generalization capability for NP-hard problems (e.g., graph-based combinatorial optimization, etc.) beyond any pre-specified problem size. It is used (Bello et al., 2016) as an encoder-decoder based learning technique, where an encoder is used to produce the embedding of the starting location and locations to be visited, whereas the decoder predicts the sequence of the visited locations. Since this concept is implemented in a static context, further investigation is followed to tackle the dynamic case of the problem (demand for visiting a location changes to zero after the traveling salesman visits it), which consists of an RNN decoder coupled with an attention mechanism (Kool et al., 2018). Using this concept, studies (Peng et al., 2020; Zhang et al., 2020) propose a model based on (graph) attention layers, which is invariant to the order of the input elements, thereby providing a solution that can be generalized to problems defined on general graphs.

# 3 Deep reinforcement learning in transportation research

## 3.1 Autonomous driving

Autonomous driving (AD) is a multi-disciplinary problem that consists of several individual tasks. These tasks involve: sensing the surroundings, perception of the situation, decision-making based on the perception, strategic planning for the execution of the selected action, and finally execution of the selected action by the control system (Talpaert et al., 2019; Kiran et al., 2020). Usually, these tasks are divided into two levels – higher level (HL) and lower level (LL). HL tasks involve the decision-making based on the reasoning of the surrounding environment and lower level (LL) task is in the control system to achieve the actual execution of the decision (Mirchevska et al., 2018; Chen et al., 2018). The applications of DRL based algorithms produce promising results in both levels.

### 3.1.1 Lane change behavior

Based on the sensory input from the surrounding environment, the autonomous agent needs to decide on whether it is appropriate to change the lane or keep the lane; this is a crucial HL decision to make when it comes to AD. Researchers have attempted to apply various DRL algorithms to create appropriate and safe lane changing strategies for AD (Table 1). The objectives of these studies are mainly focused on devising an optimal and safe lane changing strategy in a connected environment. Almost every paper designs the system as a single agent setting considering only the ego vehicle (the vehicle under consideration). However, there are few exceptions, such as Shalev et al. (2016), Yi (2018), and Chen et al. (2018) consider a multi-agent system, Wang et al. (2019a) design the environment as cooperative, Jiang et al. (2019) and Wang et al. (2020)'s settings perceive the system as both adversarial and cooperative from the perspective of surrounding vehicles. To deal with variable sized input which is more practical in lane change behavior, Huegle et al. (2019) incorporate a new DRL architecture – Deep Sets (Zaheer et al., 2017) and demonstrate that this approach can outperform CNN and RNN based DRL.

The design of the state space is roughly the same for all studies – position, speed, relative positions and relative speeds of ego and surrounding vehicles, distance and gap between vehicles, etc. Depending on the number of vehicles considered in the state space, the dimension can be up to 20. Although a low dimensional state space is desirable due to its easy implementation ability, oftentimes the state dimension becomes large to make it comprehensive enough to make better decisions (Chen et al., 2018). For HL decision-making, the action space needs to have at least three essential actions – turn to the left lane, turn to the right lane, and keep the current lane. Most of the studies only considered these three lateral actions in their DRL training, while the longitudinal actions – speed change by acceleration, deceleration, or keep the current speed – are achieved by some additional rule-based models. Nonetheless, we find some studies that train their DRL agents on both lateral and longitudinal actions in their DRL application (Table 1).

Safety, comfort, and efficiency are the three fundamental requirements for AD. The current practice uses different weighted attributing factors for each of these criteria in the reward signal. The common factors are – velocity maximization, collision avoidance, lane change completion, safe distance keeping, etc. Apart from these, some additional factors, such as cooperation among surrounding vehicles (Wang et al., 2019a), discouragement of near-crash actions (Wang et al., 2020; Bai et al., 2019), and avoidance of unnecessary lane change (Alizadeh et al., 2019; Chen et al., 2018; Min et al., 2018; Makantasis et al 2019; Hoel et al., 2019) are also considered.

HL decision-making alone cannot guarantee a collision-free trajectory without a transfer mechanism for this policy to be executed by LL motion control (Makantasis et al., 2019). For example, HL can decide to turn to the left lane; but in order to turn, to what extent the steering angle needs to be changed is to be determined by LL. As a result, several studies attempt to combine both the HL decision-making process and LL motion control for lane changing behavior in a hierarchical DRL architecture (Shi et al., 2019; Y. Chen et al., 2019; Duan et al., 2020). The main difference is that they introduce some LL control mechanism after a decision has been made in the HL. Wang et al. (2018a, 2019b) trained a quadratic Q-function approximator from the sensory input to decide on the yaw acceleration (rate of the rotational speed with respect to the vertical axis) to attain a lateral control directly.

### 3.1.2 Motion control in AD

Lower level (LL) control module of autonomous driving is mainly concerned with the execution of the planned trajectory or the decision taken at the higher level (HL). Vehicle control classically has been achieved with predictive control approaches such as Model Predictive Control (Paden et al., 2016). Recently, researchers are trying to develop learning-based LL control modules for AD. The actions in the control module involve the longitudinal and lateral adjustments by changing the acceleration and steering angle of the vehicle. Since both actions are continuous in nature, most of the studies utilize policy-based algorithms in an actor-critic architecture (Table 1). For motion control, the design of the state space should be more specific. Besides the vehicle specific and relative information, most studies consider the relative positions of the front and rear axles, and the current steering angle. Additionally, Vinitsky et al. (2018)'s state space contains additional information on surrounding human-driven vehicles. Lin et al. (2019) design the state space differently with a continuous function featuring the gap keeping error and delayed acceleration.

The design of the reward signal has a considerable variation. While it is common to give some reward for every successful task completion and a penalty for failure, the reward signal is also reflective of some overarching goals of the study itself. For instance, Buechel et al. (2018) design the reward signal with the longitudinal velocity difference to focus more on longitudinal motion, Vinitsky et al. (2018) try to maximize total throughput from a bottleneck, Folkers et al. (2019)'s goal is to explore a parking lot without facing obstacles. For AD motion control, another recent approach is end-to-end autonomous driving. End-to-end driving policies refer to the direct derivation of control signals from the raw image pixel as input feature recorded by the onboard camera. Studies that use this technique are listed in Table 1.

### 3.1.3 Miscellaneous tasks in AD

Besides lane changing behavior and motion control of AD, several other task-specific DRL applications exist in the literature. They are – intersection navigation, highway ramp merging, and car-following model. We also find one paper on pedestrian safety. In Table 2, we summarize all the papers in these categories. Here, we highlight the main elements of DRL formulation such as methods, state, actions, and rewards. Interested readers can find more detail of DRL applications in autonomous driving in two recent review papers Kiran et al. (2020) and Talpaert et al. (2019).

**Table 1.** Categorized list of references of DRL applications in the Lane change behavior and Motion control of AD.

| | | Lane change behavior | | Motion control | |
|---|---|---|---|---|---|
| | | **Only lateral decision** | **Both lateral and longitudinal decisions** | **Conversion of HL decision to LL control** | **End-to-end driving** |
| DQN and its variants | DQN | Mirchevska et al., 2018; Feng et al., 2019; Jiang et al., 2019; Wang et al., 2019a; Wang et al., 2019c; Chen et al., 2018 (Asynchronous); Alizadeh et al., 2019; Li and Czarnecki, 2019 (multi-objective) | Wolf et al., 2018; Min et al., 2018; Fayjie et al., 2018; Li and Czarnecki, 2019 | Lee et al., 2019 (HL); Chen et al., 2019b | |
| | Double DQN | | Zhang et al., 2019; Hoel et al. 2018; Nageshrao et., 2019; Makantasis et al., 2019 | | |
| | Dueling DQN | | Bai et al., 2019 | | |
| Actor-critic architecture | DDPG | An and Jung, 2019; Yi, 2018 | | Paxton et al., 2017; Buechel et al., 2018; Lin et al., 2019; Bejar et al., 2019; Lee et al., 2019 (LL); Wang et al., 2018b | Yu et al., 2018; Sallab et al., 2016, 2017 |
| | PPO | | Ye et al., 2020 | Folkers et al., 2019 | |
| | TRPO | | | Vinitsky et al., 2018 | |
| Other types | IRL | | Wang et al., 2020; You et al., 2019 | | |
| | HRL | Shi et al., 2019; Chen et al., 2019a; Duan et al., 2020 | Shalev et al., 2016; Nosrati et al., 2018 | Xu et al., 2018 | |
| | Deep QL | Wang et al., 2019b; Wang et al., 2018a (deep QL with normalized advantage function by Gu et al., 2016) | Mukadam et al., 2017 (with Q-masking) | Paxton et al., 2017 (HL) (algorithm by Mnih et al., 2013) | |
| | AlphaGo | | Hoel et al., 2019 | | |
| | Policy-based DRL | | | | Aradi et al., 2018 |

**Table 2.** Summary of DRL applications in miscellaneous tasks involved in AD.

| Tasks | Reference | Methods | State | Action | Reward |
|---|---|---|---|---|---|
| Car following model | Zhu et al., 2018 | DDPG | Following vehicle's speed, spacing and velocity difference | Acceleration | Disparity between simulated and observed speed and spacing |
| | Zhu et al., 2019 | DDPG | Following vehicle's speed, spacing and velocity difference | Acceleration | Function of time to collision, headway, and acceleration change |
| | Wu et al., 2019a | TRPO | State of charge, distance, speed of leader and follower (EV) | Acceleration | Function of distance and electricity consumption |
| | Qu et al., 2020 | DDPG | Speed, gap and the relative speed with the leader | Acceleration | Function of speed and time gap |
| | Bacchiani et al., 2019 | A3C (multi-agent) | Visual (space, obstacle, path) and numerical (speed, target speed, elapsed time ratio, distance to goal) | Acceleration, brake or maintain the same speed | Numerical reward for success and penalty for collision |
| Intersection navigation | Isele et al., 2018 | DQN | Image snapshot indicating heading angle and velocity (unsignalized intersection) | Wait, move forward slowly, and go | Numerical reward for success and penalty for collision |
| | Zhou et al., 2019 | DDPG | Vehicle and signal specific information | Acceleration | Function of speed, gap and predicted arrival time at the intersection |
| | Kashihara, 2017 | Deep QL | Image of roadway intersection (highway junction) | Moving up, down, left, and right | Numerical reward for success and penalty for collision |
| Highway ramp merging | Wang and Chan, 2017 | DQN with LSTM | Speeds and positions of ego, gap front and gap back vehicle | Acceleration and steering | Function of ego vehicle's acceleration, steering angle, speed, and the distance to its surrounding vehicles |
| | Wang and Chan, 2018 | Deep QL | Speeds and positions of ego, gap front and gap back vehicle | Acceleration | Function of ego vehicle's acceleration, steering angle, speed, and the distance to its surrounding vehicles |
| | Nishi et al., 2019 | Passive actor-critic | Speeds and positions of ego and gap back vehicle | Acceleration | Derived from value function and control dynamics due to action |
| Safety specific | Chae et al., 2017 | DQN (with prioritized Experience replay) | Relative position of the obstacle (pedestrian) and vehicle's speed | No brake, weak brake, mid brake, strong brake | Combination of two penalties - one for early brake and another for actual collision |

## 3.2 Adaptive traffic signal control

A fully adaptive TSC must be able to make decisions based on the real-time traffic condition at an intersection. This makes the adaptive TSC a complicated sequential decision-making problem where the solution algorithm needs to deal with a huge search space. The applications of DRL based methodologies have produced promising results in both a single intersection and coordinated intersections for a large transportation network. Especially, DQN along with its numerous variants has been successfully adopted to design adaptive TSC. The inherent advantage of DQN being integrated with CNN allows the state to be represented as a stack of images or image-like representation. Researchers are leveraging this advantage to design the state space of the TSC problem. Besides DQN, few utilize actor-critic based architecture, such as DDPG (Canas, 2017), A2C (Coşkun et al., 2018; Chu et al., 2019), and PPO (Lin et al., 2018). In some instances, LSTM is integrated with policy network to capture the partially observable feature of the environment (Shi and Chen, 2018; Chu et al., 2019) and continuous motion of the vehicle (Choe et al., 2018).

For adaptive TSC, the objective of DRL is to train an AI agent to adjust the traffic signal timings dynamically based on the real-time environment of the intersection. Focusing on this objective, the reward function is usually designed considering features like vehicle waiting or delay time, queue length, or discharge. The most common approach to design the state space is to present the vehicular positions, speeds, and current signal phase in an image-like grid representation detected by preinstalled traffic sensors. Retrieving the state information directly from the raw pixel of the intersection snapshot is also investigated (Table 3). The design of action space depends on the intersection configuration and simulation complexity considered in the problem formulation. With no provisions for left-turn, right-turn, and U-turn, the action space for the two-phase system consists of only two actions – 1) green on east-west traffic; 2) green on north-south traffic. Four-phase intersections have no such restrictions; hence the adopted action space is more comprehensive. Continuous action space is also considered by some researchers (Table 3). In this case, instead of changing the phase at every time step, the phase duration is updated.

For adaptive TSC in a large scale transportation network, intersections should be coordinated properly to reduce overall waiting time. Multi-agent DRL has the potential to succeed in such an environment where every intersection can be considered as an independent and decentralized entity where the decision has been made by one agent. Since decisions are taken basically at the individual intersection level, the design of state and action space is similar for both single and coordinated intersections. The difference is how to combine the reward to ensure coordination and cooperation which is usually achieved by a centralized global optimization. Literature also contains several single-agent attempts to achieve adaptive TSC for coordinated intersections. Table 3 categorizes the existing DRL based TSC literature based on the variation in state and action space, reward signal, problem setup, and algorithm used. We recommend a recent paper by Haydari et al. (2020) for a comprehensive review of both RL and DRL applications in TSC.

**Table 3.** Summary of DRL applications in adaptive TSC.

| Attributes | | Single intersection | Coordinated intersections |
|---|---|---|---|
| State space | Raw pixel from the intersection snapshot | Mousavi et al., 2017; Garg et al., 2018 | |
| | Vehicular information from the detector | Gao et al., 2017; Genders and Razavi, 2016; Li et al., 2016; Ha-li and Ke, 2017; Liang et al., 2019; Muresan et al., 2019; Wan and Hwang, 2018; Choe et al., 2018; Coşkun et al., 2018; Shabestary and Abdulhai, 2018; Shi and Chen, 2018; Calvo and Dusparic, 2018 | Van der Pol and Oliehoek, 2016; Gong et al., 2019; Tan et al., 2019; Zhang et al., 2019; Lin et al., 2018; Liu et al., 2017; Shi and Chen, 2018; Calvo and Dusparic, 2018; Ge et al., 2019; Liu et al., 2018 |
| | Others | Canas, 2017 | Chu et al., 2019; Canas, 2017 |
| Action space | Two-phase intersection | Li et al. (2016), Gao et al. (2017) and Mousavi et al. (2017) | Tan et al., 2019 |
| | More than two-phase | Genders and Razavi, 2016, Wan and Hwang, 2018; Choe et al., 2018; Coşkun et al., 2018; Shabestary and Abdulhai, 2018; Calvo and Dusparic, 2018 | Van der Pol and Oliehoek, 2016; Gong et al., 2019; Chu et al., 2019; Zhang et al., 2019; Lin et al., 2018; Liu et al., 2017; Shi and Chen, 2018; Calvo and Dusparic, 2018; Ge et al., 2019; Liu et al., 2018 |
| | Phase update | Liang et al. 2019; Canas, 2017 | Canas, 2017 |
| Reward | Waiting time/delay | Genders and Razavi, 2016; Gao et al., 2017; Mousavi et al., 2017; Liang et al. (2019); Wan and Hwang, 2018; Shabestary and Abdulhai, 2018; Choe et al., 2018 | Gong et al., 2019; Shi and Chen, 2018; Calvo and Dusparic, 2018; Liu et al., 2017; Ge et al., 2019 |
| | Queue length/discharge | Muresan et al., 2019 | Lin et al., 2018; Ge et al., 2019 |
| | Combination of both | Li et al., 2016 | Van der Pol and Oliehoek, 2016; Tan et al., 2019; Chu et al., 2019; Zhang et al., 2019 |
| | Others | Canas, 2017 | Canas, 2017; |
| Algorithm | DQN | Gao et al., 2017; Li et al., 2016; Liang et al., 2019; Wan and Hwang, 2018; Shabestary and Abdulhai, 2018; Choe et al., 2018; Coşkun et al., 2018; Genders and Razavi, (2016), | Van der Pol and Oliehoek, 2016; Gong et al., 2019; Ge et al., 2019 |
| | DRL with CNN | Ha-li and Ke, 2017; Muresan et al., 2019 | |
| | Actor-critic based | Coşkun et al., 2018; Canas, 2017 | Chu et al., 2019; Canas, 2017; Lin et al., 2018 |

| | Attributes | Single intersection | Coordinated intersections |
|---|---|---|---|
| Problem setup | Others | Mousavi et al., 2017; Shi and Chen, 2018; Calvo and Dusparic, 2018; Garg et al., 2018; | Zhang et al., 2019; Shi and Chen, 2018; Calvo and Dusparic, 2018; Liu et al., 2017 |
| | Single-agent setting | Gao et al., 2017; Genders and Razavi, 2016; Li et al., 2016; Ha-li and Ke, 2017; Liang et al., 2019; Muresan et al., 2019; Wan and Hwang, 2018; Shabestary and Abdulhai, 2018; Choe et al., 2018; Coşkun et al., 2018 | Canas, 2017; Liu et al., 2018; Lin et al., 2018 |
| | Multi-agent setting | | Van der Pol and Oliehoek, 2016; Gong et al., 2019; Chu et al., 2019; Zhang et al., 2019; Ge et al., 2019; Liu et al., 2017; Shi and Chen, 2018; Calvo and Dusparic, 2018 |

## 3.3 Energy efficient driving

The emergence of electric vehicle (EV) creates an avenue for reduced energy consumption and low emission driving. Pure electric vehicles often fall short of operational ranges and are still comparatively expensive to produce and develop (Liessner et al., 2018a). Hybrid electric vehicles (HEV) offers a good alternative by combining the benefits of the conventional internal combustion engine (ICE) and electric motors (EM) that has the potential to reduce the emission as well as solve the low range problem associated with EV. One of the important components of HEV is the energy management system (EMS) that essentially decides on optimal control strategies combing both electricity and fossil fuel targeting to achieve the best possible energy efficiency. Based on the configurations, HEV can be classified into three main classes: series, parallel, power split (Sabri et al., 2016). Although their operating modes and power management systems are different, a proper energy management system (EMS) is necessary for every configuration to optimally manage the power split at a regular interval. Lately, DRL techniques have been producing some promising results in this decision-making process. In order to deal with continuous state and action space, many adopt actor-critic based DDPG algorithm to design the EMS of HEV (Liessner et al., 2018a, 2018b; Li et al., 2019a, 2019b; Lian et al., 2020; Wang et al., 2019d; Tan et al., 2019; Wu et al., 2019b). Our review also reveals that many researchers have successfully applied DQN (Qi et al., 2017; Hu et al., 2018; Wu et al., 2018a), dueling DQN (Qi et al., 2019; Li et al., 2018), double Q-learning (Wang et al., 2019e; Han et al., 2019) to the EMS of HEV.

The recent DRL applications in the EMS of HEV are focused on overcoming the limitations involved in traditional rule-based and optimization-based approaches (Wu et al., 2019b), such as the requirement of expert knowledge, comprehensive driving cycles, extensive road information, accurate prediction, etc. The goal here is to train an agent to make autonomous decisions on the optimal split between electricity and fuel solely based on vehicle interactions with the roadway and vehicle dynamics. Although some researchers are skeptical about imposing any prior rule while training DRL agents, several studies show that the overall learning process and performance can be further enhanced by imposing rules (Wang et al., 2019d, 2019e; Lian et al., 2020). To attain the desired objective of autonomous EMS, the existing studies design the reward signal based on the energy consumptions and savings by keeping track of the fuel consumption rate and state-of-charge (SOC) of the batteries (Qi et al., 2017, 2019; Liessner et al., 2018a, 2018b, 2019; Li et al., 2018, 2019a, 2019b: Lian et al., 2020; Tan et al., 2019; Zhao et al., 2018; Chaoui et al., 2018; Han et al., 2019; Hu et al., 2018; Wu et al., 2018a, 2019b). Considering an HEV based last-mile delivery system, Wang et al. (2019d, 2019e)'s design of the reward signal additionally considers the change in overall trip distance.

For EMS, the most important information the agent needs to know is vehicle dynamics (velocity and acceleration) and the energy state of the vehicle (power demand and SOC). Every study in the literature considers these two types of information. In addition to the current velocity, Tan et al. (2019) and Wu et al. (2019b) keep track of velocities from several previous states. A more granular level of state representation is adopted by Liessner et al. (2018, 2019). The authors consider wheel rotation and torque, gear configuration, battery temperature, and derating effect. Apart from the vehicle specific internal information, several studies incorporate external information on distance to destination (Qi et al., 2019), traveled distance, and expected future travel distance (Wang et al., 2019d, 2019e; Wu et al., 2019b) and roadway conditions such as terrain and slope (Li et al., 2019b, 2020).

The design of the action space is mainly focused on how to optimize the energy usages between the power sources. Some try to attain this by only changing the power supply from ICE (Qi et al., 2019; Hu et al., 2018; Han et al., 2019; Wu et al., 2018a, 2019b; Lian et al., 2020), few other consider changing energy output only from EM (Liessner et al., 2018a, 2019; Zhao et al., 2018). Another way can be keeping a balance among multiple inputs of the same power source (Li et al., 2019a, 2019b). Training is mainly done in simulated platforms; however, some utilize real-world trip data for training and testing (Qi et al., 2017, 2019; Wang et al., 2019d, 2019e). Hyperparameter optimization is another important phase in the DRL approach. For the EMS of HEV, Liessner et al. (2019) propose a model-based hyperparameter optimization technique. Application of DRL in numerous hybrid vehicles, such as series-parallel plug-in hybrid electric bus (Wu et al., 2019b), power-split hybrid electric bus (Wu et al., 2018a), hybrid electric tracked vehicle (Han et al., 2019), multiple battery based EV (Chaoui et al., 2018), hybrid electric bus (Tan et al., 2019), plug-in HEV (Hu et al., 2018), series HEV (Li et al., 2018) show better results than their baselines.

## 3.4 Maritime freight transportation

Most of the DRL applications in maritime transportation is in Autonomous Ship (AS) driving, more specifically in path following and collision avoidance of AS. Given the dynamics and complexity of an AS system, it is argued that existing analytical control methods (model predictive control) are not very robust to be appropriate for practical applications (Zhao et al., 2019). On the other hand, since a strength of DRL is its ability to tackle dynamic and complex systems, DRL has been considered and shown success in path following of AS. Several researchers consider both the collision-free path and path following problems simultaneously in their DRL models (Table 4).

The DRL based path following problem aims to design a dynamic system that can be implemented to describe the dynamic behavior of the vehicle. A wide range of control methods for ship collision avoidance has been conducted extensively. Most of them can be divided into two classes: model-based and model-free methods. DRL is found to be used in model-free methods. Among them, several approaches only focus on static obstacles and do not consider dynamic obstacles i.e., environmental disturbances. Although DRL has the potential to be applied in the port management, its application is very rare in the literature. We find a study (Shen et al., 2017) focusing on the quay and yard crane scheduling problem where DQN is applied for the crane scheduling in Ningbo Port. Table 4 summarizes the above-reviewed studies in terms of application area, method, state, action, and reward.

**Table 4.** Summary of DRL applications in the maritime freight literature.

| Application area | | Reference | Method | State | Action | Reward |
|---|---|---|---|---|---|---|
| Path following | | Martinsen and Lekkas, 2018 | Actor-critic | Cross-track error, course and heading to the path, surge, sway, yaw rate, and derivatives of the path | Rudder angle | Negative cross-track error |
| | | Rejaili and Figueiredo, 2018 | DQN and DDPG | Distance and longitudinal axis of AS to the guideline, horizontal and vertical speed, and angular velocity | Rudder angles and propeller rotations | Deviation from the guideline and the speed setpoint |
| | | Woo et al., 2019 | DDPG | Position | Course angle | Deviation from route |
| Path following and collision avoidance | Static object | Layek et al., 2017 | DDPG | Position, orientation, and actual turning rate | Turning rate | Angle between ship and obstacles |
| | | Amendola et al., 2019 | DQN | Distance between AS and centerline of channel, course over ground, rate-of-turn, and last rudder level | Rudder angle | Proximity from AS to its destination, avoiding collision, and deviation of vessel from center |
| | | Sawada, 2019 | PPO | Position, AS's breadth, length, and speed | Heading and rudder angle | Safe passing distance and deviation from route |
| | | Shen et al., 2019 | DQN | Distance from obstacles and other ship | Rudder angle | Obstacle avoidance |
| | | Zhang et al., 2019 | DQN | Speed, ship course, position of obstacle and target point, distance between AS and target point, and the distance between AS and obstacle | Heading angle | Approach to the target point and obstacle avoidance |
| | Static and dynamic object | Cheng and Zhang, 2018 | DQN | Position, heading, surge, sway and the angular rate of vessel, and obstacle's position | Rudder angle | Obstacle avoidance and target approach |
| | | Cheng-bo et al., 2019 | DQN | Current position, speed and course of AS, distance from target and obstacles, and obstacle position | Heading angle | Proximity from AS to its destination, avoiding collision, and deviation of vessel from center |
| | | Etemad et al., 2020 | DQN | Position | Heading angle | Approach to the target point and obstacle avoidance |
| | | Zhao and Roh, 2019 | Multi-agent policy-based DRL | Current position, velocity and length of AS, distance between current position and destination, and relative angle between ship and the destination | Rudder angle | Proximity from AS to its destination, avoiding collision, and drift |
| | | Wang et al., 2019g | DQN | Speed, position of obstacles and destination, and distance of vessel from obstacles and destination | Rudder angle | Approach to the target point, obstacle avoidance, and deviation from route |
| | | Zhao et al., 2019 | Policy-based DRL | Current position, velocity and length of AS, distance between current position and destination, and relative angle between ship and the destination | Rudder angle | Proximity from AS to its destination, avoiding collision, and drift |
| | | Guo et al., 2020 | Actor critic | Longitude and latitude | Heading angle, speed | Approach to the target point and obstacle avoidance |
| Port management | | Shen et al., 2017 | DQN | Container loading condition in every time step | Scheduling container | Availability, reshuffling, and yard crane shifting |

## 3.5 Route optimization in transportation

The use and adaptation of DRL for routing optimization has been multifold, spanning many types of problems from the more basic traveling salesman problems (Bello et al., 2016; Khalil et al., 2017; Kool et al., 2018) to more complicated multi-vehicle routing problems (Nazari et al., 2018; Kullman et al., 2019; Zhang et al., 2020; Peng et al., 2020; James et al., 2019; Chen et al., 2019c; Chen et al., 2019d; Balaji et al., 2019), and to taxi dispatching and routing decisions (Oda and Tachibana, 2018; Oda and Wong, 2018; Shi et al., 2019; Al-Abbasi et al., 2019; Singh et al., 2019; Kullman et al., 2020; Holler et al., 2020; Wen et al., 2017). These DRL based VRP studies propose a mix of techniques: 1) attention models based on an encoder-decoder architecture (Zhang et al., 2020; Peng et al., 2020; Nazari et al., 2018); 2) a graph embedding network to represent the policy that captures the property of a node in the context of its graph neighborhood (Khalil et al., 2017); 3) an Atari-fied representation of the environment (Kullman et al., 2019). Given the intimate connection of traveling salesman problems to multi-vehicle routing problems, below we categorize our review into two groups - solving vehicle routing problems (VRP) including both freight delivery problems and emerging mobility services intended for people transportation (i.e., shared taxi route optimization).

The complexity of routing problems is augmented when extended to multiple routes, with time constraints, and with pickups and deliveries. Several recent efforts have appeared in the literature, on both freight and passenger sides. For freight delivery problems, Nazari et al. (2018) consider a parameterized stochastic policy to solve a VRP with limited vehicle capacity. The authors apply a policy gradient DRL algorithm to optimize the parameters of the stochastic policy. Chen et al. (2019c) use multi-agent RL to train a courier dispatch policy to deal with goods pickups with prescribed pickup time windows. To maintaining the state-action space at a controllable size, the RL is decentralized with each courier modeled as an agent. However, the drawback of a decentralized approach is the compromise of interactions among couriers undertaking different pickup tasks. Similar to Chen et al. (2019c), Yu et al. (2019) opt for a distributed neural optimization strategy to solve a pickup and delivery problem with the capacity constraint for vehicle and time constraint for request delivery. The authors adopt a graph embedded pointer network to progressively develop a complete tour for each vehicle. In a similar vein, Chen et al. (2019d) investigate a heterogeneous fleet of vehicles and drones for same-day delivery service and Balaji et al. (2019) solve an on-demand delivery driver model, which is essentially a VRP variant (stochastic dynamic VRP).

DRL based solution approaches to solving shared taxi route optimization problems are also garnering growing attentions lately (Oda and Tachibana, 2018; Oda and Wong, 2018; Shi et al., 2019; Al-Abbasi et al., 2019; Singh et al., 2019; Kullman et al., 2020; Holler et al., 2020; Wen et al., 2017). The most common practice in the literature is considering the system as a distributed dispatching framework (i.e., learning is independent for each vehicle) without coordination with other vehicles in its vicinity (Oda and Joe-Wong, 2018; Al-Abbasi et al., 2019; Singh et al., 2019). Furthermore, several studies consider anticipated future demand while designing their state space, which proves to have played a very significant role in securing optimal dispatch policies by proactively repositioning vehicles to mitigate the spatial imbalance (Wen et al., 2017; Al-Abbasi et al., 2019; Holler et al., 2019; Kullman et al., 2020). The goal is to move idle vehicles proactively in parallel with regular assignment actions. However, the formulation of reward to achieve that "goal" is often varied in the literature, such as saving in waiting time (Wen et al., 2017), maximizing revenue from request assignments (Holler et al., 2019; Kullman et al., 2020), and minimizing idle/en-route time and fuel usages (Al-Abbasi et al., 2019).

## 3.6 Rail transportation

Our review of the literature reveals that DRL has produced promising results in Train Timetable Rescheduling (TTR) (Ning et al., 2019; Obara et al., 2018; Wang et al., 2019f; Yang et al., 2019), Automatic Train Operations (ATO) (Zhou and Song, 2018; Zhou et al., 2020; Zhu et al., 2017), and Train Shunting Operations (TSO) (Peer et al., 2018). TTR problem involves finding a feasible timetable of a train either by re-routing, re-ordering, re-timing, or canceling in case of any uncertain disturbances resulting in an equipment/system failure along the railway line. As the TTR problem is NP-hard, finding an optimal solution by traditional optimization method is time-consuming. On the contrary, DRL is a quick solver and can be successfully used in the TTR problem. Another field of application of DRL is the ATO system in which the velocity and trajectory decision of a high-speed train can be determined. ATO system must tackle some uncertain situations (i.e., changing trip time, longer waiting time for passenger boarding) of the operation that requires a model-free optimization technique. TSO includes matching the timetable of the incoming and outgoing trains as well as scheduling the maintenance and cleaning activities in the shunting yard. In this complex environment, DRL is applied successfully in the Dutch railway (Peer et al., 2018).

Three types of DRL models are used in the literature of rail transportation management. The first one is DQN, which is used in solving the TTR problem (Ning et al., 2019; Obara et al., 2018), ATO system (Zhu et al., 2017), and TSO (Peer et al., 2018). DDPG is used by Yang et al. (2019), Zhou and Song (2018) and Zhou et al. (2020). Monte Carlo tree search is used in Wang et al. (2019f). Although most of the existing studies adopt simulated environments for their experiment, Peer et al. (2018), Yang et al. (2019) and Zhou et al. (2020) resort to real train track information of Dutch railways, Shanghai Metro Line and Yizhuang Line of the Beijing Subway, respectively.

TTR aims to address the recovery of train operation order in reordering and retiming strategies during disturbances with minimum delay (Ning et al., 2019; Wang et al., 2019f), the dissatisfaction of the passenger (Obara et al., 2018), and reduce energy consumption (Yang et al., 2019). State, action, and reward used by Ning et al. (2019) and Wang et al. (2019f) are the same. They define actual arrival and departure times as state, reorder of the departure sequences as action, and negative average total delay as reward. Obara et al. (2018) describe the delay situation in a graph environment and consider graph deformation in the action space. They consider delay, stoppage, driving time, frequency, and connection in their reward function. As the aim of the study of Yang et al. (2019) is to reduce energy consumption, they use recovered energy and the negative of traction energy as the reward function.

The DRL based ATO system aims to minimize energy consumption (Zhou and Song, 2018; Zhu et al., 2017) constrained by delay (Zhu et al., 2017), punctuality, and riding comfort (Zhou et al., 2020). Zhou and Song (2018) and Zhou et al. (2020) define speed and the position of the train as the state. The magnitude of acceleration and deceleration is considered as action. However, their reward function is different. Zhou and Song (2018) consider energy consumption in their reward function whereas Zhou et al., (2020) use both energy consumption and delay as their reward function. To minimize the profile tracking error and energy consumption, Zhu et al. (2017) design their reward function with the deviation of the speed from the target speed profile to take a decision on either acceleration or deceleration. In their state space, they include speed, train position, the relative position from the front train, strength of the rail wireless network, and a binary indicator that specifies whether the train starts using the wireless network or not. Peer et al. (2018) introduce arrival train condition and departure train condition in the shunting yard as the state and parking and departure decision of each track as the action aiming to minimize the error in parking and departure from the shunting yard.

## 3.7  Traffic management by road control

DRL can be highly beneficial to traffic management systems by automating numerous road control strategies, such as speed limit control, ramp metering, lane pricing, etc., which in turn enhances highway safety and mitigates congestion. Variable speed limits (VSL) control offers an adaptive and flexible means to revamp traffic conditions, increase safety, and reduce emissions. These applications adopt total travel time or total delay as the intended objective of a VSL control system. Therefore, existing studies define the reward function to account for the total travel time, vehicular emission (Zhu and Ukkusuri, 2014; Wu et al., 2018b), and total delay (Nezafat, 2019). Besides, Wu et al. (2018b) introduce another objective which attempts to reduce the crash probability.

For VSL, state space is usually designed with traffic density and discretized congestion levels (Nezafat, 2019). Wu et al. (2018b) suggest incorporating information such as the numbers of lanes at the immediate upstream, mainline, and on-ramp of merge area and the occupancy rate which can be collected exploiting loop detectors. Researchers use actor-critic based policy gradient algorithms such as – DDPG (Wu et al., 2018b), A3C (Nezafat, 2019) due to the continuous nature of the action space which is determining the speed limit within a road section. The results of the studies clearly indicate that DRL can outperform other state-of-the-art feedback-control based solution and Q-learning based solution when it comes to reduction in travel time and emission (Nezafat, 2019; Wu et al., 2018b)

DRL is also applied in other traffic management systems such as ramp metering and lane pricing. DRL outperforms current state-of-the-art ramp metering policy ALINEA (Papageorgiou et al. 1997) by achieving a precise adaptive highway ramp metering without any model calibration (Belletti et al., 2017). Authors use a partial differential equation to simulate the environment and integrate deep neural networks to approximate policies form the REINFORCE algorithm (Schulman et al., 2015). Pandey et al. (2019) apply DRL for the dynamic pricing for managed lanes. Considering the environment as a partially observable Markov decision process, their policy gradient approach is able to change the tolls dynamically. In a multi-objective formulation, the results outperform existing heuristics-based solution in terms of both revenue generation and travel time minimization.

# 4 Synthesized discussions

## 4.1 Applicability

With the recent advancement in DRL methodology and applications and computation power, it is possible to design and solve many sequential decision-making problems as a DRL problem. However, to design a problem in the context of DRL, one must understand the basic nature of the problem as well as DRL. A DRL algorithm tries to optimize a problem by improving the policies that result from the agent interacting with the environment. Therefore, any problem where the solution can be improved by trial-and-error (i.e., incorporating the feedback from the environment from each trial to the next) would be a good candidate for the DRL application. Furthermore, problems that give more importance to completing a full task rather than periodical success in some intermediate steps can be designed as a DRL problem, with the concept of delayed reward. End-to-end autonomous driving is an example of such problems where a collision before the end of the journey overshadows the earlier success in the lane change or intermediate steps. A simulated environment is necessary to train a DRL agent. Due to the obvious reason, a DRL agent cannot be trained in the real world. But without proper training, a real-world application is out of the question. Therefore, a simulated environment needs to be sophisticated enough to capture important details from the real world. Finally, sequential decision-making needs to be formulated in a mathematical framework. MDP is the most suitable framework for the fully observable environment. However, DRL can also be applied in the partially observable environment with a partially observable Markov decision process formulation by incorporating RNN with a function approximator (Heess et al., 2015; Hausknecht et al., 2015).

It is difficult to say which DRL algorithm is perfect for a problem. However, depending on the nature of the problems, we can identify some candidate algorithms. For example, problems where states can be incorporated as a stack of images, the algorithms embedded with CNN as a function approximator are the most appropriate, i.e., DQN. The state-of-the-art of DQN is the combination of both double DQN and dueling DQN along with the prioritized experience replay. Therefore, problems like adaptive signal control, end-to-end autonomous driving can adopt double dueling DQN with prioritized experience replay. Both policy-based and actor-critic based algorithms can deal with continuous action space. The actor-critic architecture is preferred when both the optimal policy and its value are expected. Hence, algorithms like DDPG and PPO can be applied for the EMS for hybrid or electric vehicles and motion control for autonomous driving. One interesting area of exploration can be a comparison between the performance of DDPG and PPO. Asynchronous algorithms like A3C is also based on actor-critic architecture and can accelerate learning speed by parallel processing. Although numerical experiment in Atari gaming platform shows that PPO can produce a better solution than A3C, still A3C can be considered for transportation application for its high learning speed. Based on the existing literature, an illustration of the applications of various DRL algorithms and extensions in seven transportation domains is presented in Fig. 4. The left column of this figure lists some popular DRL algorithms, and the right column lists some extensions popularly used in DRL applications. The algorithm acronyms are mentioned in section 2. The middle column indicates application areas. The arrow widths in this figure are proportional to the frequency of applications in the existing literature.

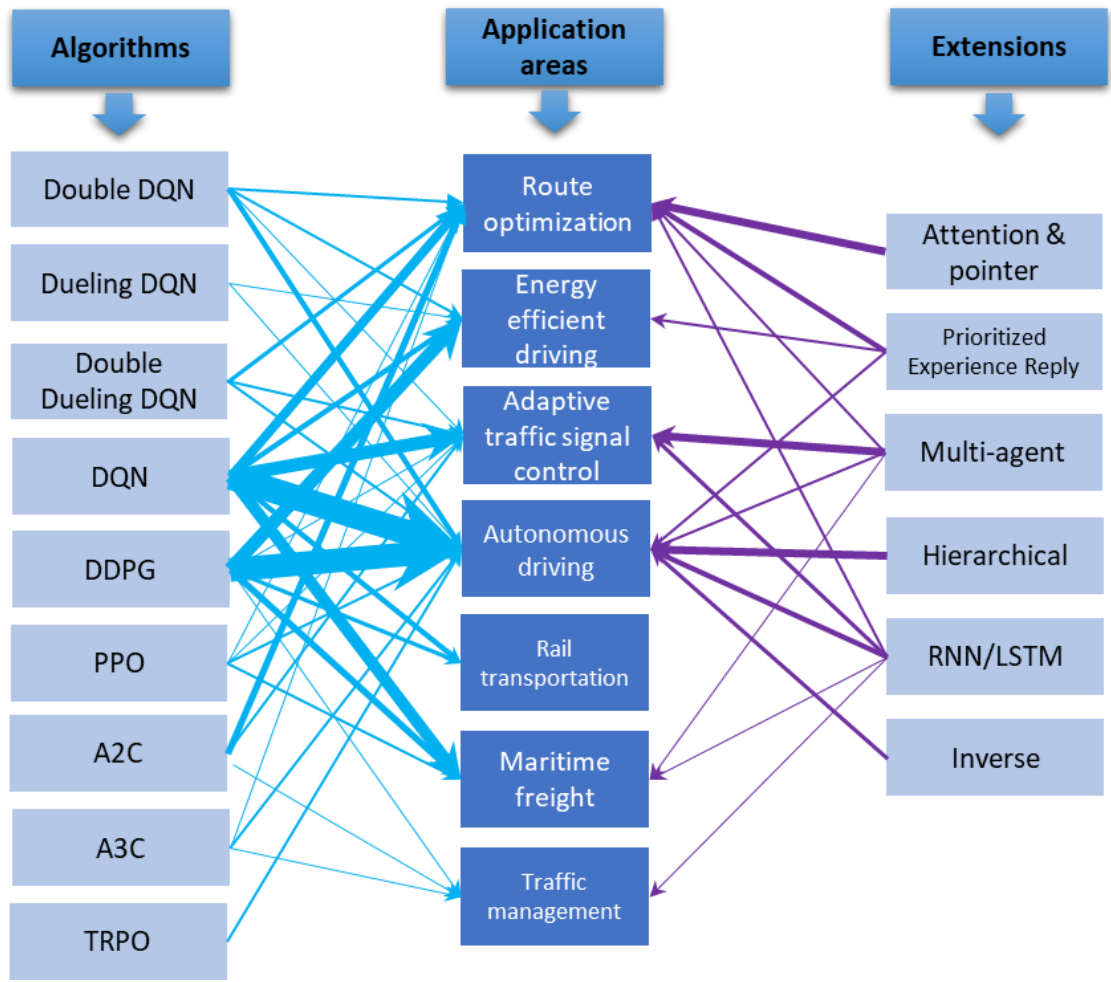

**Figure 4.** DRL application in transportation research.

To formulate an autonomous and connected environment, multi-agent DRL can be considered. In the road network, one can design both cooperative and adversarial intentions of the neighboring vehicles with a multi-agent system. Multi-agent DRL algorithms can also be considered for adaptive TSC in coordinated intersections, and trajectory optimizations of autonomous car and ship in dynamic environments. When the decisions can be decomposed into multiple layers, we should consider hierarchical approaches of DRL. For instance, if the action space can be divided into two levels – "what to do" and "how to do" – then HRL can make the overall learning and implementation process very efficient. For instance, in the routing and scheduling problem, which neighborhood moves to select can be considered as higher level decisions and how to adopt this selected neighborhood move can be designed as lower level sub-policies. In the autonomous driving, the decision on the action can be considered as a higher level problem; on the other hand, the actual execution of the decision can be considered as a lower level control problem.

## 4.2 Strengths and shortcomings of DRL

*Strengths*: Through some extensive training of DRL algorithms, some extraordinary performance has been reported in autonomous decision-making tasks (Minh et al., 2015; Silver et al., 2016, 2017). The complex decision-making tasks involved in transportation operations are perfect candidates for the application of DRL algorithms. From the extensive review presented in this paper, it is quite evident that DRL can produce results that surpass the quality of existing benchmarks and yet it takes only a split second to produce solutions that would usually take considerably longer times by classical heuristics and metaheuristics.

While interacting with the environment, a DRL algorithm gathers experiences of countless unique situations. Therefore, it has the capability of producing good solutions even with completely new problem instances. The advantage of RL being implementable in uncharted territory makes it superior to other classical optimization techniques. Incorporating DL with RL has boosted the capability of DRL in terms of scalability and applicability. Especially, considerably superior results have been obtained by DRL algorithms in large state and action spaces (Minh et al., 2015) and even in continuous action spaces (Lillicrap et al., 2016; Schulman et al., 2017).

*Shortcomings:* Training of DRL algorithms can be unstable. Especially, DQN is often criticized for unstable training. The DQN algorithm involves a target network that is updated at regular intervals. At each update, the values of different parameters could experience unstable jumps (Van der Pol and Oliehoek, 2016). Therefore, it may be difficult to see a clear trend of learning improvement during the training process. Deep learning itself is an emerging science. There still remain lots of unanswered questions about its interpretability (Chakraborty et al., 2017; Zhang and Zhu, 2018). The lack of interpretability makes it difficult to see the level of contribution from each of the state information in the final solution. Furthermore, in general DRL algorithms suffer from a lack of reproducibility (Hoffman et al., 2020). These inherent shortcomings also affect their applications in transportation.

DRL involves updating the weights of an artificial neural network. Training a neural network involves hundreds of thousands of iterations over an enormous amount of data. With the advent of simulated environments, researchers often turn to synthetic data for the training of the neural network. This extensive training calls for very high computational power. With state-of-the-art GPU, parallel processing paves an efficient way of updating the weights in the neural network. However, GPUs are often very expensive. Although the deep learning platforms like Tensorflow, PyTorch have built-in functionality to optimize the usages between CPU and GPU, it often requires a tradeoff between the computational efficiency and financial capability.

Hyperparameter tuning is one of the essential tasks of the DRL algorithms. Every DRL algorithm involves a certain number of hyperparameters to be tuned. This tuning process is a very time consuming and even frustrating task. Tuning involves a grid search of all the hyperparameters. It thus would require a vast number of combinations and permutations with the hyperparameters. The training of a neural network is usually very slow; for some problem instances, it can take in the order of days. Therefore, one can imagine the colossal amount of time it involves to fine-tune a DRL algorithm.

## 4.3 Issues and future directions for DRL applications to transportation

### 4.3.1 Common issues and research directions

With the advent of DRL algorithms in recent years, there is no doubt that the potential of DRL applications in transportation is far-reaching. In many transportation fields to which DRL is adopted, results have been reported to outperform known benchmarks. Despite solution superiority, however, we cannot conclude that DRL is mature enough to be applied to all real-world applications. As shown in Section 3, existing DRL transportation research is mainly based on simulated platforms with synthetic data. Although some use historical data to train the agent, still, without explicitly testing in real-world scenarios DRL cannot be applied in practice with absolute confidence. No matter how many scenarios are exposed to the DRL agents in the simulation, these solution strategies will hardly be perceived as reliable or safe enough without testing them in the real world. Another significant question is the transferability of learned policies to the real world. For this, only limited attempts have been made for autonomous driving (Xu et al., 2018; Chalaki et al., 2019; Chen et al., 2018). The transferability issue remains an open question and warrants further attention.

For the same type of problems, the dimension of state can vary significantly. This makes it difficult to come up with a proper scale of comparison. A common platform would be needed to evaluate the effectiveness of different DRL algorithm variations for the same problem type. This would facilitate observing, tracking, comparing, and ultimately understanding the methodological strengths of different algorithm variations. Furthermore, in the literature DRL algorithms are often applied in tailored and simplified environments. In future research, the performance of DRL solutions should be evaluated in more generic and comprehensive environments to identify the best solutions that are robust to different environments.

For DRL applications in TSC and AD, researchers often use some statistical distributions to simulate traffic flow in a road network. The existing literature has neither reached an agreement on the appropriateness of any single distribution nor made comparison to illustrate the most appropriate one. We also note a debate between the preference of a rule-based learning approach and a completely rule-independent learning approach in DRL based EMS applications. Further investigations will be needed to bring forth concrete recommendations on rule-based vs. rule-independent learning for real-world implementation, which is true not only for EMS but any other DRL based applications. In addition, as modern vehicles are equipped with advanced technologies such as radar, GPS, lidar, and camera, researchers may consider leveraging these onboard technologies to collect real-time information to support DRL applications in vehicle level operation decision and control.

### 4.3.2 Application-specific issues and research directions

Three issues are present in AD. First, some actions (acceleration, speed, etc.) are often designed as a discretized and stepwise fashion. However, using policy based or actor-critic based algorithms, these actions can be considered as continuous which would be more realistic. Second, in DRL based AD literature, sensing and perceiving the behavior of surrounding vehicles are two essential tasks before making a decision. However, the existing literature often fails to distinguish human-driven and autonomous vehicles whose behaviors will be different. DRL based AD studies need to take this issue into account in the environment design. Another major drawback in DRL based AD literature is the narrow study focus or scope. Most of the existing studies mainly focus on one or two specific aspects of AD, such as lane changing behavior, motion control, and collision avoidance. However, to achieve a safe, reliable, and efficient AD, the DRL agent has to make decisions on all the aspects that build up the entire AD. Although the sum of successes of individual aspects leads to the overall success, these individual successes need to be summed in an efficient manner, which is yet to be accomplished.

For adaptive TSC, we have seen some significant advantages of DRL based methods over traditional approaches. However, besides the issue of lack of real-world testing, DRL based TSC still needs to address some shortcomings. The design of state space should accommodate the acceleration and deceleration of incoming traffic in the yellow and red phase for both single and coordinated intersections. This shortcoming is identified by one of the earliest DRL studies on adaptive TSC (Genders and Razavi, 2016) but yet to be addressed. The grid-based traffic representation considers a Boolean-valued vector where a value 1 indicates the presence of a vehicle and a value 0 indicates the absence of a vehicle, such assumption may not very practical for real-world application. The DRL agent optimizes the overall traffic waiting time in the intersection by dynamically adjusting the signal timings. This goal may not achieve uniform prioritization of every vehicle waiting in the intersection. Attention to the individual level is still lagging in the current design of problem formulation.

In the case of DRL based EMS, the main objective is to achieve vehicle fuel economy. However, the reward signal is usually designed based on the power supply from the engine. The relationship between fuel economy and engine power is complex and most of the papers lack a proper explanation for this (Hu et al., 2018). The success of EMS of any HEVs depends not only on the state and the dynamics of the vehicle and the surrounding environment, but also other factors such as expected future information including trip distance and trip time, average vehicle speed attainable during a trip, and driver's driving behavior that are not readily available to the agent and can vary during the trip. A few recent studies attempt to take those "other factors" into account (Wang et al., 2019d, 2019e; Wu et al. 2019b). But more extensive investigation will still be needed. Additionally, further research efforts should be made for DRL based EMS for plug-in HEV, considering the increasing popularity and market penetration of plug-in HEV.

As has been reviewed, considerable progress has been made in solving combinatorial optimization problems like TSP and VRP with DRL, which leads to novel approaches to taxi routing and dispatching and parcel delivery problems. A few areas still call for more attention, for example, dealing with conflicting interests in reward formulation (i.e., minimizing the unserved requests, waiting time, total travel time of users, and the vehicles' total trip distance) and maintaining coordination among the fleet of vehicles. Furthermore, the success of DRL in TSP and VRP has the potential to expand to more specific applications like pickup-and-delivery, crowdshipping, and vehicle platooning problems.

For rail transportation, the issue of scalability has not been paid much attention in DRL applications. Also, due to the lack of well-defined benchmarks, it is difficult to assess the quality of DRL solutions. In future endeavors, researchers should pay more attention to comparing the performance of their DRL solutions with solutions from existing heuristics. For maritime freight transportation, existing DRL based solution strategies do not take environmental disturbance issues into account. Yet ocean current and wind speed are very important factors and should be considered in the state space design. DRL may be further used to obtain energy efficient navigation.

# 5 Available resources for DRL research

Implementation of DRL is not a straightforward process, it requires a reliable platform where DRL algorithms can be developed, implemented, documented, and tested. Fortunately, there exist some built-in platforms to facilitate the development of new DRL algorithms or to apply existing algorithms to a variety of environments. OpenAI Baselines (Dhariwal et al., 2017) provide a set of high-quality implementations for the reproducibility of different DRL algorithms. Stable Baselines (Hill et al, 2017) is a big improvement upon OpenAI Baselines, featuring a platform for all algorithms. Moreover, TensorForce (Kuhnle, 2017) is a framework for DRL built around DL library Tensorflow with several algorithm implementations. Tensorflow Agents (Guadarrama et al., 2018) is a versatile RL platform developed by the TensorFlow library where an agent takes charge of two main responsibilities: 1) defining a policy to interact with the environment, and 2) how to learn that policy from collected experience. KerasRL (Plappert, 2016) implements some state-of-the-art RL algorithms in Python and seamlessly integrates with the DL library Keras. Additionally, rlpyt (Stooke and Abbeel, 2019) implements deep Q-learning, policy gradients, and Q-value policy gradients algorithm for small to medium scale research in DRL.

**Table 5.** A summary of some properties of the aforementioned DRL frameworks.

| Platform | Available DRL algorithm | Library used |
|---|---|---|
| OpenAI Baselines | A2C, Actor-Critic with Experience Replay, Actor-Critic using Kronecker-Factored Trust Region, DDPG, DQN, Generative Adversarial Imitation Learning (GAIL), Hindsight Experience Replay (HER), TRPO, PPO, | Tensorflow |
| Stable Baselines | A2C, ACER, ACKTR, DDPG, double dueling DQN with prioritized experience replay, GAIL, HER, PPO, TRPO, Soft Actor-Critic (SAC) | Tensorflow |
| TensorForce | Dueling and double DQN, Vanilla Policy Gradient (PG), Continuous DQN (CDQN), A2C, A3C, TRPO, PPO | Tensorflow |
| Tensorflow Agents | DQN, double DQN, DDPG, Twin Delayed Deep Deterministic policy gradient algorithm (TD3), Simple Statistical Gradient-Following Algorithms for Connectionist Reinforcement Learning, PPO, SAC | Tensorflow |
| KerasRL | Dueling and double DQN, DDPG, CDQN, Cross-Entropy Method (CEM), Deep SARSA | Keras |
| rlpyt | A2C, PPO, Dueling and double DQN, DDPG, TD3, SAC | PyTorch |

# 6 Conclusion

This paper is devoted to creating a comprehensive review and synthesis of DRL applications in transportation research. Various DRL algorithms have been successfully applied across several transportation fields. In the area of autonomous driving, DRL application has brought some significant improvement in both higher level decision-making and lower level control problems. Successful applications of DRL have also been observed in adaptive traffic signal control for both single and coordinated intersections, and in traffic management systems in which various road control measures such as dynamic speed limit control, road pricing, and lane pricing are considered. Hybrid and electric vehicle's energy management system is another area where DRL algorithms have attracted considerable attention. The area of route optimization is taking advantage of the significant advancement as well in DRL based TSP and VRP research. The success of DRL further extends to maritime freight and rail applications. Based on this review, synthesized discussions on the applicability, strength, shortcomings, and issues of DRL algorithms in transportation applications are conducted, based on which future research directions are recommended. Last but not least, this review provides a documentation of available resources for actually implementing DRL. We hope that this review can serve as a useful reference to help the transportation research community better understand what have been accomplished to date and what are the potentials of DRL for transportation, and stimulate further research in this exciting and rapidly developing area.

# Acknowledgement

This work presented in this paper was funded in part by the National Science Foundation CMMI-1663411. The support is gratefully acknowledged.

# References

1. Al-Abbasi, A. O., Ghosh, A., and Aggarwal, V. (2019). Deeppool: Distributed model-free algorithm for ride-sharing using deep reinforcement learning. IEEE Transactions on Intelligent Transportation Systems, 20(12), 4714-4727.
2. Alizadeh, A., Moghadam, M., Bicer, Y., Ure, N. K., Yavas, U., and Kurtulus, C. (2019). Automated Lane Change Decision Making using Deep Reinforcement Learning in Dynamic and Uncertain Highway Environment. In 2019 IEEE Intelligent Transportation Systems Conference (ITSC) (pp. 1399-1404). IEEE.
3. Amendola, J., Tannuri, E. A., Cozman, F. G., and Reali Costa, A. H. (2019). Port Channel Navigation Subjected to Environmental Conditions Using Reinforcement Learning. In International Conference on Offshore Mechanics and Arctic Engineering (Vol. 58844, p. V07AT06A042). American Society of Mechanical Engineers.
4. An, H., and Jung, J. I. (2019). Decision-making system for lane change using deep reinforcement learning in connected and automated driving. Electronics, 8(5), 543.
5. Aradi, S., Becsi, T., and Gaspar, P. (2018). Policy gradient based reinforcement learning approach for autonomous highway driving. In 2018 IEEE Conference on Control Technology and Applications (CCTA) (pp. 670-675). IEEE.
6. Arulkumaran, K., Deisenroth, M. P., Brundage, M., and Bharath, A. A. (2017). A brief survey of deep reinforcement learning. arXiv preprint arXiv:1708.05866.
7. Bacchiani, G., Molinari, D., and andPatander, M. (2019). Microscopic traffic simulation by cooperative multi-agent deep reinforcement learning. In Proceedings of the 18th International Conference on Autonomous Agents and MultiAgent Systems (pp. 1547-1555). International Foundation for Autonomous Agents and Multiagent Systems.
8. Bahdanau, D., Cho, K., and Bengio, Y. (2014). Neural machine translation by jointly learning to align and translate. *arXiv preprint arXiv:1409.0473*.
9. Bai, Z., Shangguan, W., Cai, B., and Chai, L. (2019). Deep Reinforcement Learning Based High-level Driving Behavior Decision-making Model in Heterogeneous Traffic. In 2019 Chinese Control Conference (CCC) (pp. 8600-8605). IEEE.
10. Balaji, B., Bell-Masterson, J., Bilgin, E., Damianou, A., Garcia, P.M., Jain, A., Luo, R., Maggiar, A., Narayanaswamy, B. and Ye, C. (2019). ORL: Reinforcement Learning Benchmarks for Online Stochastic Optimization Problems. arXiv preprint arXiv:1911.10641.

11. Barto, A. G., and Mahadevan, S. (2003). Recent advances in hierarchical reinforcement learning. Discrete event dynamic systems, 13(1-2), 41-77.

12. Bejar, E., and Morán, A. (2019). Reverse parking a car-like mobile robot with deep reinforcement learning and preview control. In 2019 IEEE 9th Annual Computing and Communication Workshop and Conference (CCWC) (pp. 0377-0383). IEEE.

13. Belletti, F., Haziza, D., Gomes, G., and Bayen, A. M. (2017). Expert level control of ramp metering based on multi-task deep reinforcement learning. IEEE Transactions on Intelligent Transportation Systems, 19(4), 1198-1207.

14. Bello, I., Pham, H., Le, Q. V., Norouzi, M., and Bengio, S. (2016). Neural combinatorial optimization with reinforcement learning. arXiv preprint arXiv:1611.09940.

15. Brackstone, M., and McDonald, M. (1999). Car-following: a historical review. Transportation Research Part F: Traffic Psychology and Behaviour, 2(4), 181-196.

16. Buechel, M., and Knoll, A. (2018). Deep reinforcement learning for predictive longitudinal control of automated vehicles. In 2018 21st International Conference on Intelligent Transportation Systems (ITSC) (pp. 2391-2397). IEEE.

17. Buşoniu, L., Babuška, R., and De Schutter, B. (2010). Multi-agent reinforcement learning: An overview. In Innovations in multi-agent systems and applications-1 (pp. 183-221). Springer, Berlin, Heidelberg.

18. Chakraborty, S., Tomsett, R., Raghavendra, R., Harborne, D., Alzantot, M., Cerutti, F., Srivastava, M., Preece, A., Julier, S., Rao, R.M. and Kelley, T.D. (2017). Interpretability of deep learning models: a survey of results. In 2017 IEEE SmartWorld, Ubiquitous Intelligence and Computing, Advanced and Trusted Computed, Scalable Computing and Communications, Cloud and Big Data Computing, Internet of People and Smart City Innovation (SmartWorld/SCALCOM/UIC/ATC/CBDCom/IOP/SCI) (pp. 1-6). IEEE.

19. Chae, H., Kang, C. M., Kim, B., Kim, J., Chung, C. C., and Choi, J. W. (2017). Autonomous braking system via deep reinforcement learning. In 2017 IEEE 20th International Conference on Intelligent Transportation Systems (ITSC) (pp. 1-6). IEEE.

20. Chalaki, B., Beaver, L., Remer, B., Jang, K., Vinitsky, E., Bayen, A., and Malikopoulos, A. A. (2019). Zero-shot autonomous vehicle policy transfer: From simulation to real-world via adversarial learning. arXiv preprint arXiv:1903.05252.
44

21. Chaoui, H., Gualous, H., Boulon, L., and Kelouwani, S. (2018). Deep reinforcement learning energy management system for multiple battery based electric vehicles. In 2018 IEEE Vehicle Power and Propulsion Conference (VPPC) (pp. 1-6). IEEE.
22. Chen, C., Qian, J., Yao, H., Luo, J., Zhang, H., and Liu, W. (2018). Towards comprehensive maneuver decisions for lane change using reinforcement learning.
23. Chen, I. M., Zhao, C., and Chan, C. Y. (2019b). A Deep Reinforcement Learning-Based Approach to Intelligent Powertrain Control for Automated Vehicles. In 2019 IEEE Intelligent Transportation Systems Conference (ITSC) (pp. 2620-2625). IEEE.
24. Chen, X., Ulmer, M. W., and Thomas, B. W. (2019d). Deep Q-Learning for Same-Day Delivery with a Heterogeneous Fleet of Vehicles and Drones. arXiv preprint arXiv:1910.11901.
25. Chen, Y., Dong, C., Palanisamy, P., Mudalige, P., Muelling, K., and Dolan, J. M. (2019a). Attention-based Hierarchical Deep Reinforcement Learning for Lane Change Behaviors in Autonomous Driving. In Proceedings of the IEEE Conference on Computer Vision and Pattern Recognition Workshops (pp. 0-0).
26. Chen, Y., Qian, Y., Yao, Y., Wu, Z., Li, R., Zhou, Y., Hu, H. and Xu, Y.(2019c). Can Sophisticated Dispatching Strategy Acquired by Reinforcement Learning?. In Proceedings of the 18th International Conference on Autonomous Agents and MultiAgent Systems (pp. 1395-1403). International Foundation for Autonomous Agents and Multiagent Systems.
27. Cheng, Y., and Zhang, W. (2018). Concise deep reinforcement learning obstacle avoidance for underactuated unmanned marine vessels. Neurocomputing, 272, 63-73.
28. Cheng-bo, W. A. N. G., Xin-yu, Z. H. A. N. G., Jia-wei, Z. H. A. N. G., Zhi-guo, D. I. N. G., and Lan-xuan, A. N. (2019). Navigation behavioural decision-making of MASS based on deep reinforcement learning and artificial potential field. In Journal of Physics: Conference Series (Vol. 1357, No. 1, p. 012026). IOP Publishing.
29. Chu, T., Wang, J., Codecà, L., and Li, Z. (2019). Multi-agent deep reinforcement learning for large-scale traffic signal control. IEEE Transactions on Intelligent Transportation Systems.
30. Dayan, P., and Hinton, G. E. (1993). Feudal reinforcement learning. In Advances in neural information processing systems (pp. 271-278).
31. Deisenroth, M. P., Neumann, G., and Peters, J. (2013). A survey on policy search for robotics. now publishers.
32. Dhariwal, P., Hesse, C., Klimov, O., Nichol, A., Plappert, M., Radford, A., Schulman, J., Sidor, S., Wu, Y. and Zhokhov, P., (2017). Openai baselines.

33. Duan, J., Li, S. E., Guan, Y., Sun, Q., and Cheng, B. (2020). Hierarchical reinforcement learning for self-driving decision-making without reliance on labelled driving data. IET Intelligent Transport Systems.

34. Etemad, M., Zare, N., Sarvmaili, M., Soares, A., Machado, B. B., and Matwin, S. (2020). Using Deep Reinforcement Learning Methods for Autonomous Vessels in 2D Environments. arXiv preprint arXiv:2003.10249.

35. Fayjie, A. R., Hossain, S., Oualid, D., and Lee, D. J. (2018). Driverless car: Autonomous driving using deep reinforcement learning in urban environment. In 2018 15th International Conference on Ubiquitous Robots (UR) (pp. 896-901). IEEE.

36. Feng, X., Hu, J., Huo, Y., and Zhang, Y. (2019). Autonomous Lane Change Decision Making Using Different Deep Reinforcement Learning Methods. In CICTP 2019 (pp. 5563-5575).

37. Ferdowsi, A., Challita, U., Saad, W., and Mandayam, N. B. (2018). Robust deep reinforcement learning for security and safety in autonomous vehicle systems. In 2018 21st International Conference on Intelligent Transportation Systems (ITSC) (pp. 307-312). IEEE.

38. Folkers, A., Rick, M., and Büskens, C. (2019). Controlling an Autonomous Vehicle with Deep Reinforcement Learning. In 2019 IEEE Intelligent Vehicles Symposium (IV) (pp. 2025-2031). IEEE.

39. Gao, J., Shen, Y., Liu, J., Ito, M., and Shiratori, N. (2017). Adaptive traffic signal control: Deep reinforcement learning algorithm with experience replay and target network. arXiv preprint arXiv:1705.02755.

40. Garg, D., Chli, M., and Vogiatzis, G. (2018). Deep reinforcement learning for autonomous traffic light control. In 2018 3rd IEEE International Conference on Intelligent Transportation Engineering (ICITE) (pp. 214-218). IEEE.

41. Genders, W., and Razavi, S. (2016). Using a deep reinforcement learning agent for traffic signal control. arXiv preprint arXiv:1611.01142.

42. Gong, Y., Abdel-Aty, M., Cai, Q., and Rahman, M. S. (2019). Decentralized network level adaptive signal control by multi-agent deep reinforcement learning. Transportation Research Interdisciplinary Perspectives, 1, 100020.

43. Grondman, I., Busoniu, L., Lopes, G. A., and Babuska, R. (2012). A survey of actor-critic reinforcement learning: Standard and natural policy gradients. IEEE Transactions on Systems, Man, and Cybernetics, Part C (Applications and Reviews), 42(6), 1291-1307.

44. Gu, S., Lillicrap, T., Sutskever, I., and Levine, S. (2016). Continuous deep q-learning with model-based acceleration. In International Conference on Machine Learning (pp. 2829-2838).
45. Guadarrama, S., Korattikara, A., Ramirez, O., Castro, P., Holly, E., Fishman, S., Wang, K., Gonina, E., Harris, C., Vanhoucke, V. and Brevdo, E., 2018. TF-Agents: A library for reinforcement learning in tensorflow.
46. Guo, S., Zhang, X., Zheng, Y., and Du, Y. (2020). An Autonomous Path Planning Model for Unmanned Ships Based on Deep Reinforcement Learning. Sensors, 20(2), 426.
47. Ha-li, P., and Ke, D. (2017). An intersection signal control method based on deep reinforcement learning. In 2017 10th International Conference on Intelligent Computation Technology and Automation (ICICTA) (pp. 344-348). IEEE.
48. Han, X., He, H., Wu, J., Peng, J., and Li, Y. (2019). Energy management based on reinforcement learning with double deep Q-learning for a hybrid electric tracked vehicle. Applied Energy, 254, 113708.
49. Harmon, M. E., Baird III, L. C., and Klopf, A. H. (1995). Advantage updating applied to a differential game. In Advances in neural information processing systems (pp. 353-360).
50. Hausknecht, M., and Stone, P. (2015). Deep recurrent q-learning for partially observable mdps. In 2015 AAAI Fall Symposium Series.
51. Haydari, A., and Yilmaz, Y. (2020). Deep Reinforcement Learning for Intelligent Transportation Systems: A Survey. arXiv preprint arXiv:2005.00935.
52. Heess, N., Hunt, J. J., Lillicrap, T. P., and Silver, D. (2015). Memory-based control with recurrent neural networks. arXiv preprint arXiv:1512.04455.
53. Hernandez-Leal, P., Kartal, B., and Taylor, M. E. (2019). A survey and critique of multiagent deep reinforcement learning. Autonomous Agents and Multi-Agent Systems, 33(6), 750-797.
54. Hester, T., Vecerik, M., Pietquin, O., Lanctot, M., Schaul, T., Piot, B., Dan, H., Quan, J., Sendonaris, A., Osband, I., Dulac-Arnold, G., Agapiou, J., Leibo, J. Z., Gruslys, A. (2018). Deep q-learning from demonstrations. In Thirty-Second AAAI Conference on Artificial Intelligence.
55. Hill, A., Raffin, A., Ernestus, M., Gleave, A., Traore, R., Dhariwal, P., Hesse, C., Klimov, O., Nichol, A., Plappert, M. and Radford, A., 2018. Stable baselines. GitHub repository.
56. Hochreiter, S., and Schmidhuber, J. (1997). Long short-term memory. Neural computation, 9(8), 1735-1780.

57. Hoel, C. J., Driggs-Campbell, K., Wolff, K., Laine, L., and Kochenderfer, M. (2019). Combining planning and deep reinforcement learning in tactical decision making for autonomous driving. IEEE Transactions on Intelligent Vehicles.

58. Hoel, C. J., Wolff, K., and Laine, L. (2018). Automated speed and lane change decision making using deep reinforcement learning. In 2018 21st International Conference on Intelligent Transportation Systems (ITSC) (pp. 2148-2155). IEEE.

59. Hoffman, M., Shahriari, B., Aslanides, J., Barth-Maron, G., Behbahani, F., Norman, T., Abdolmaleki, A., Cassirer, A., Yang, F., Baumli, K. and Henderson, S. (2020). Acme: A Research Framework for Distributed Reinforcement Learning. arXiv preprint arXiv:2006.00979.

60. Holler, J., Vuorio, R., Qin, Z., Tang, X., Jiao, Y., Jin, T., Singh, S., Wang, C. and Ye, J. (2019). Deep Reinforcement Learning for Multi-Driver Vehicle Dispatching and Repositioning Problem. arXiv preprint arXiv:1911.11260.

61. Hu, Y., Li, W., Xu, K., Zahid, T., Qin, F., and Li, C. (2018). Energy management strategy for a hybrid electric vehicle based on deep reinforcement learning. Applied Sciences, 8(2), 187.

62. Huegle, M., Kalweit, G., Mirchevska, B., Werling, M., and Boedecker, J. (2019). Dynamic Input for Deep Reinforcement Learning in Autonomous Driving. arXiv preprint arXiv:1907.10994.

63. Huegle, M., Kalweit, G., Werling, M., and Boedecker, J. (2019). Dynamic Interaction-Aware Scene Understanding for Reinforcement Learning in Autonomous Driving. arXiv preprint arXiv:1909.13582.

64. Ioffe, S., and Szegedy, C. (2015). Batch normalization: Accelerating deep network training by reducing internal covariate shift. arXiv preprint arXiv:1502.03167.

65. Isele, D., Rahimi, R., Cosgun, A., Subramanian, K., and Fujimura, K. (2018). Navigating occluded intersections with autonomous vehicles using deep reinforcement learning. In 2018 IEEE International Conference on Robotics and Automation (ICRA) (pp. 2034-2039). IEEE.

66. James, J. Q., Yu, W., and Gu, J. (2019). Online vehicle routing with neural combinatorial optimization and deep reinforcement learning. IEEE Transactions on Intelligent Transportation Systems, 20(10), 3806-3817.

67. Jiang, S., Chen, J., and Shen, M. (2019). An Interactive Lane Change Decision Making Model With Deep Reinforcement Learning. In 2019 7th International Conference on Control, Mechatronics and Automation (ICCMA) (pp. 370-376). IEEE.

68. Kakade, S., and Langford, J. (2002). Approximately optimal approximate reinforcement learning. In ICML (Vol. 2, pp. 267-274).

69. Kashihara, K. (2017). Deep Q learning for traffic simulation in autonomous driving at a highway junction. In 2017 IEEE International Conference on Systems, Man, and Cybernetics (SMC) (pp. 984-988). IEEE.

70. Khalil, E., Dai, H., Zhang, Y., Dilkina, B., and Song, L. (2017). Learning combinatorial optimization algorithms over graphs. In Advances in Neural Information Processing Systems (pp. 6348-6358).

71. Kiran, B. R., Sobh, I., Talpaert, V., Mannion, P., Sallab, A. A. A., Yogamani, S., and Pérez, P. (2020). Deep reinforcement learning for autonomous driving: A survey. arXiv preprint arXiv:2002.00444.

72. Konda, V. R., and Tsitsiklis, J. N. (2000). Actor-critic algorithms. In Advances in neural information processing systems (pp. 1008-1014).

73. Kool, W., Hoof, H. V., and Welling, M. (2018). Attention solves your TSP, approximately. Statistics, 1050, 22.

74. Koutník, J., Cuccu, G., Schmidhuber, J., and Gomez, F. (2013). Evolving large-scale neural networks for vision-based reinforcement learning. In Proceedings of the 15th annual conference on Genetic and evolutionary computation (pp. 1061-1068).

75. Kuhnle, A., Schaarschmidt, M., and Fricke, K. (2017). Tensorforce: a tensorflow library for applied reinforcement learning. Web page.

76. Kulkarni, T. D., Narasimhan, K., Saeedi, A., and Tenenbaum, J. (2016). Hierarchical deep reinforcement learning: Integrating temporal abstraction and intrinsic motivation. In Advances in neural information processing systems (pp. 3675-3683).

77. Kullman, N. D., Mendoza, J. E., Cousineau, M., and Goodson, J. C. (2019). Atari-fying the Vehicle Routing Problem with Stochastic Service Requests. arXiv preprint arXiv:1911.05922.

78. Kullman, N., Cousineau, M., Goodson, J., and Mendoza, J. (2020). Dynamic Ridehailing with Electric Vehicles.

79. Laurent, G. J., Matignon, L., and Fort-Piat, L. (2011). The world of independent learners is not Markovian. International Journal of Knowledge-based and Intelligent Engineering Systems, 15(1), 55-64.

80. Layek, A., Vien, N. A., and Chung, T. (2017). Deep reinforcement learning algorithms for steering an underactuated ship. In 2017 IEEE International Conference on Multisensor Fusion and Integration for Intelligent Systems (MFI) (pp. 602-607). IEEE

81. LeCun, Y., Bengio, Y., and Hinton, G. (2015). Deep learning. nature, 521(7553), 436-444.

82. Lee, J., Balakrishnan, A., Gaurav, A., Czarnecki, K., and Sedwards, S. (2019). Wisemove: A framework for safe deep reinforcement learning for autonomous driving. arXiv preprint arXiv:1902.04118.
83. Li, C., and Czarnecki, K. (2019). Urban driving with multi-objective deep reinforcement learning. In Proceedings of the 18th International Conference on Autonomous Agents and MultiAgent Systems (pp. 359-367). International Foundation for Autonomous Agents and Multiagent Systems.
84. Li, Y. (2017). Deep reinforcement learning: An overview. arXiv preprint arXiv:1701.07274.
85. Li, L., Lv, Y., and Wang, F. Y. (2016). Traffic signal timing via deep reinforcement learning. IEEE/CAA Journal of Automatica Sinica, 3(3), 247-254.
86. Li, Y., He, H., Khajepour, A., Wang, H., and Peng, J. (2019). Energy management for a power-split hybrid electric bus via deep reinforcement learning with terrain information. Applied Energy, 255, 113762.
87. Li, Y., He, H., Peng, J., and Wang, H. (2019). Deep Reinforcement Learning-Based Energy Management for a Series Hybrid Electric Vehicle Enabled by History Cumulative Trip Information. IEEE Transactions on Vehicular Technology, 68(8), 7416-7430.
88. Li, Y., He, H., Peng, J., and Wu, J. (2018). Energy Management Strategy for a Series Hybrid Electric Vehicle Using Improved Deep Q-network Learning Algorithm with Prioritized Replay. DEStech Transactions on Environment, Energy and Earth Sciences, (iceee).
89. Lian, R., Peng, J., Wu, Y., Tan, H., and Zhang, H. (2020). Rule-interposing deep reinforcement learning based energy management strategy for power-split hybrid electric vehicle. Energy, 117297.
90. Liang, X., Du, X., Wang, G., and Han, Z. (2019). A deep reinforcement learning network for traffic light cycle control. IEEE Transactions on Vehicular Technology, 68(2), 1243-1253.
91. Liessner, R., Dietermann, A. M., and Bäker, B. (2018). Safe deep reinforcement learning hybrid electric vehicle energy management. In International Conference on Agents and Artificial Intelligence (pp. 161-181). Springer, Cham.
92. Liessner, R., Schmitt, J., Dietermann, A., and Bäker, B. (2019). Hyperparameter Optimization for Deep Reinforcement Learning in Vehicle Energy Management. In ICAART 2019.
93. Liessner, R., Schroer, C., Dietermann, A. M., and Bäker, B. (2018). Deep Reinforcement Learning for Advanced Energy Management of Hybrid Electric Vehicles. In ICAART (2) (pp. 61-72).

94. Lillicrap, T.P., Hunt, J.J., Pritzel, A., Heess, N., Erez, T., Tassa, Y., Silver, D. and Wierstra, D. (2015). Continuous control with deep reinforcement learning. arXiv preprint arXiv:1509.02971.
95. Lin, Y., McPhee, J., and Azad, N. L. (2019). Longitudinal dynamic versus kinematic models for car-following control using deep reinforcement learning. In 2019 IEEE Intelligent Transportation Systems Conference (ITSC) (pp. 1504-1510). IEEE.
96. Littman, M. L. (1994). Markov games as a framework for multi-agent reinforcement learning. In Proceedings of the 11th international conference on machine learning (pp. 157–163). New Brunswick, NJ, USA
97. Littman, M. L. (2001). Value-function reinforcement learning in Markov games. Cognitive Systems Research, 2(1), 55–66
98. Makantasis, K., Kontorinaki, M., and Nikolos, I. (2019). A deep reinforcement learning driving policy for autonomous road vehicles. arXiv preprint arXiv:1905.09046.Martinsen, A. B., and Lekkas, A. M. (2018). Curved path following with deep reinforcement learning: Results from three vessel models. In OCEANS 2018 MTS/IEEE Charleston (pp. 1-8). IEEE.
99. Min, K., and Kim, H. (2018). Deep q learning based high level driving policy determination. In 2018 IEEE Intelligent Vehicles Symposium (IV) (pp. 226-231). IEEE.
100. Mirchevska, B., Pek, C., Werling, M., Althoff, M., and Boedecker, J. (2018). High-level decision making for safe and reasonable autonomous lane changing using reinforcement learning. In 2018 21st International Conference on Intelligent Transportation Systems (ITSC) (pp. 2156-2162). IEEE.
101. Mnih, V., Badia, A.P., Mirza, M., Graves, A., Lillicrap, T., Harley, T., Silver, D. and Kavukcuoglu, K. (2016). Asynchronous methods for deep reinforcement learning. In International conference on machine learning (pp. 1928-1937).
102. Mnih, V., Kavukcuoglu, K., Silver, D., Rusu, A.A., Veness, J., Bellemare, M.G., Graves, A., Riedmiller, M., Fidjeland, A.K., Ostrovski, G. and Petersen, S. (2015). Human-level control through deep reinforcement learning. Nature, 518(7540), 529-533.
103. Mnih, V., Kavukcuoglu, K., Silver, D., Graves, A., Antonoglou, I., Wierstra, D., and Riedmiller, M. (2013). Playing atari with deep reinforcement learning. arXiv preprint arXiv:1312.5602.
104. andMousavi, S. S., Schukat, M., and Howley, E. (2017). Traffic light control using deep policy-gradient and value-function-based reinforcement learning. IET Intelligent Transport Systems, 11(7), 417-423.

105. Mukadam, M., Cosgun, A., Nakhaei, A., and Fujimura, K. (2017). Tactical decision making for lane changing with deep reinforcement learning.
106. Muresan, M., Fu, L., and Pan, G. (2019). Adaptive traffic signal control with deep reinforcement learning an exploratory investigation. arXiv preprint arXiv:1901.00960.
107. Nageshrao, S., Tseng, H. E., and Filev, D. (2019). Autonomous highway driving using deep reinforcement learning. In 2019 IEEE International Conference on Systems, Man and Cybernetics (SMC) (pp. 2326-2331). IEEE.
108. Nazari, M., Oroojlooy, A., Snyder, L., and Takác, M. (2018). Reinforcement learning for solving the vehicle routing problem. In Advances in Neural Information Processing Systems (pp. 9839-9849).
109. Nezafat, R. V. (2019). Deep Reinforcement Learning Approach for Lagrangian Control: Improving Freeway Bottleneck Throughput Via Variable Speed Limit.
110. Ng, A. Y., and Russell, S. J. (2000). Algorithms for inverse reinforcement learning. In Icml (Vol. 1, p. 2).
111. Ning, L., Li, Y., Zhou, M., Song, H., and Dong, H. (2019). A Deep Reinforcement Learning Approach to High-speed Train Timetable Rescheduling under Disturbances. In 2019 IEEE Intelligent Transportation Systems Conference (ITSC) (pp. 3469-3474). IEEE.
112. Nishi, T., Doshi, P., and Prokhorov, D. (2019). Merging in congested freeway traffic using multipolicy decision making and passive actor-critic learning. IEEE Transactions on Intelligent Vehicles, 4(2), 287-297.
113. Nosrati, M.S., Abolfathi, E.A., Elmahgiubi, M., Yadmellat, P., Luo, J., Zhang, Y., Yao, H., Zhang, H. and Jamil, A. (2018). Towards practical hierarchical reinforcement learning for multi-lane autonomous driving.
114. Nowé, A., Vrancx, P., and De Hauwere, Y. M. (2012). Game theory and multi-agent reinforcement learning. In Reinforcement Learning (pp. 441-470). Springer, Berlin, Heidelberg.
115. Obara, M., Kashiyama, T., and Sekimoto, Y. (2018). Deep Reinforcement Learning Approach for Train Rescheduling Utilizing Graph Theory. In 2018 IEEE International Conference on Big Data (Big Data) (pp. 4525-4533). IEEE
116. Oda, T., and Joe-Wong, C. (2018). MOVI: A model-free approach to dynamic fleet management. In IEEE INFOCOM 2018-IEEE Conference on Computer Communications (pp. 2708-2716). IEEE.

117. Oda, T., and Tachibana, Y. (2018). Distributed fleet control with maximum entropy deep reinforcement learning.
118. Pandey, V., and Boyles, S. D. (2018). Dynamic pricing for managed lanes with multiple entrances and exits. Transportation Research Part C: Emerging Technologies, 96, 304-320.
119. Papageorgiou, M., Hadj-Salem, H., and Middelham, F. (1997). ALINEA local ramp metering: Summary of field results. Transportation research record, 1603(1), 90-98.
120. Paxton, C., Raman, V., Hager, G. D., and Kobilarov, M. (2017). Combining neural networks and tree search for task and motion planning in challenging environments. In 2017 IEEE/RSJ International Conference on Intelligent Robots and Systems (IROS) (pp. 6059-6066). IEEE.
121. Peer, E., Menkovski, V., Zhang, Y., and Lee, W. J. (2018). Shunting trains with deep reinforcement learning. In 2018 ieee international conference on systems, man, and cybernetics (smc) (pp. 3063-3068). IEEE.
122. Peng, B., Wang, J., and Zhang, Z. (2019). A Deep Reinforcement Learning Algorithm Using Dynamic Attention Model for Vehicle Routing Problems. In International Symposium on Intelligence Computation and Applications (pp. 636-650). Springer, Singapore.
123. Plappert, M. (2016). Keras-rl. GitHub Repository.
124. Qi, X., Luo, Y., Wu, G., Boriboonsomsin, K., and Barth, M. (2019). Deep reinforcement learning enabled self-learning control for energy efficient driving. Transportation Research Part C: Emerging Technologies, 99, 67-81.
125. Qi, X., Luo, Y., Wu, G., Boriboonsomsin, K., and Barth, M. J. (2017). Deep reinforcement learning-based vehicle energy efficiency autonomous learning system. In 2017 IEEE Intelligent Vehicles Symposium (IV) (pp. 1228-1233). IEEE.
126. Qu, X., Yu, Y., Zhou, M., Lin, C. T., and Wang, X. (2020). Jointly dampening traffic oscillations and improving energy consumption with electric, connected and automated vehicles: A reinforcement learning based approach. Applied Energy, 257, 114030.
127. Rejaili, R. P. A., and Figueiredo, J. M. P. (2018). Deep reinforcement learning algorithms for ship navigation in restricted waters. Mecatrone, 3(1).
128. Sallab, A. E., Abdou, M., Perot, E., and Yogamani, S. (2016). End-to-end deep reinforcement learning for lane keeping assist. arXiv preprint arXiv:1612.04340.
129. Sallab, A. E., Abdou, M., Perot, E., and Yogamani, S. (2017). Deep reinforcement learning framework for autonomous driving. Electronic Imaging, 2017(19), 70-76.

130. Sawada, R. (2019). Automatic Collision Avoidance Using Deep Reinforcement Learning with Grid Sensor. In Symposium on Intelligent and Evolutionary Systems (pp. 17-32). Springer, Cham.

131. Schaul, T., Quan, J., Antonoglou, I., and Silver, D. (2015). Prioritized experience replay. arXiv preprint arXiv:1511.05952.

132. Schester, L., and Ortiz, L. E. (2019). Longitudinal Position Control for Highway On-Ramp Merging: A Multi-Agent Approach to Automated Driving. In 2019 IEEE Intelligent Transportation Systems Conference (ITSC) (pp. 3461-3468). IEEE.

133. Schulman, J., Levine, S., Abbeel, P., Jordan, M., and Moritz, P. (2015). Trust region policy optimization. In International conference on machine learning (pp. 1889-1897).

134. Schulman, J., Moritz, P., Levine, S., Jordan, M., and Abbeel, P. (2015). High-dimensional continuous control using generalized advantage estimation. arXiv preprint arXiv:1506.02438.

135. Schulman, J., Wolski, F., Dhariwal, P., Radford, A., and Klimov, O. (2017). Proximal policy optimization algorithms. arXiv preprint arXiv:1707.06347.

136. Shalev-Shwartz, S., Shammah, S., and Shashua, A. (2016). Safe, multi-agent, reinforcement learning for autonomous driving. arXiv preprint arXiv:1610.03295.

137. Sharifzadeh, S., Chiotellis, I., Triebel, R., and Cremers, D. (2016). Learning to drive using inverse reinforcement learning and deep q-networks. arXiv preprint arXiv:1612.03653.

138. Shen, H., Hashimoto, H., Matsuda, A., Taniguchi, Y., Terada, D., and Guo, C. (2019). Automatic collision avoidance of multiple ships based on deep Q-learning. Applied Ocean Research, 86, 268-288.

139. Shen, Y., Zhao, N., Xia, M., and Du, X. (2017). A deep q-learning network for ship stowage planning problem. Polish Maritime Research, 24(s3), 102-109.

140. Shi, D., Ding, J., Errapotu, S. M., Yue, H., Xu, W., Zhou, X., and Pan, M. (2019). Deep $Q$-Network-Based Route Scheduling for TNC Vehicles With Passengers' Location Differential Privacy. IEEE Internet of Things Journal, 6(5), 7681-7692.

141. Shi, T., Wang, P., Cheng, X., Chan, C. Y., and Huang, D. (2019). Driving Decision and Control for Automated Lane Change Behavior based on Deep Reinforcement Learning. In 2019 IEEE Intelligent Transportation Systems Conference (ITSC) (pp. 2895-2900). IEEE.

142. Silver, D., Lever, G., Heess, N., Degris, T., Wierstra, D., and Riedmiller, M. (2014). Deterministic policy gradient algorithms.

143. Silver, D., Huang, A., Maddison, C.J., Guez, A., Sifre, L., Van Den Driessche, G., Schrittwieser, J., Antonoglou, I., Panneershelvam, V., Lanctot, M. and Dieleman, S. (2016). Mastering the game of Go with deep neural networks and tree search. nature, 529(7587), 484.
144. Silver, D., Schrittwieser, J., Simonyan, K., Antonoglou, I., Huang, A., Guez, A., Hubert, T., Baker, L., Lai, M., Bolton, A. and Chen, Y. (2017). Mastering the game of go without human knowledge. Nature, 550(7676), 354-359.
145. Singh, A., Al-Abbasi, A., and Aggarwal, V. (2019). A reinforcement learning based algorithm for multi-hop ride-sharing: Model-free approach. In Neural Information Processing Systems (Neurips) Workshop.
146. Stadie, B. C., Abbeel, P., and Sutskever, I. (2017). Third-person imitation learning. arXiv preprint arXiv:1703.01703.
147. Stooke, A., and Abbeel, P. (2019). rlpyt: A research code base for deep reinforcement learning in pytorch. arXiv preprint arXiv:1909.01500.
148. Sutton, R. S., Precup, D., and Singh, S. (1999). Between MDPs and semi-MDPs: A framework for temporal abstraction in reinforcement learning. Artificial intelligence, 112(1-2), 181-211.
149. Talpaert, V., Sobh, I., Kiran, B. R., Mannion, P., Yogamani, S., El-Sallab, A., and Perez, P. (2019). Exploring applications of deep reinforcement learning for real-world autonomous driving systems. arXiv preprint arXiv:1901.01536.
150. Tan, H., Zhang, H., Peng, J., Jiang, Z., and Wu, Y. (2019). Energy management of hybrid electric bus based on deep reinforcement learning in continuous state and action space. Energy Conversion and Management, 195, 548-560.
151. Tan, M. (1993). Multi-agent reinforcement learning: Independent vs. cooperative agents. In Proceedings of the tenth international conference on machine learning (pp. 330-337).
152. Tan, T., Bao, F., Deng, Y., Jin, A., Dai, Q., and Wang, J. (2019). Cooperative deep reinforcement learning for large-scale traffic grid signal control. IEEE transactions on cybernetics.
153. Van der Pol, E., and Oliehoek, F. A. (2016). Coordinated deep reinforcement learners for traffic light control. Proceedings of Learning, Inference and Control of Multi-Agent Systems (at NIPS 2016).
154. Van Hasselt, H. (2010). Double Q-learning. In Advances in neural information processing systems (pp. 2613-2621).
155. Van Hasselt, H., Guez, A., and Silver, D. (2016). Deep reinforcement learning with double q-learning. In Thirtieth AAAI conference on artificial intelligence.

156. Vezhnevets, A. S., Osindero, S., Schaul, T., Heess, N., Jaderberg, M., Silver, D., and Kavukcuoglu, K. (2017). Feudal networks for hierarchical reinforcement learning. In Proceedings of the 34th International Conference on Machine Learning-Volume 70 (pp. 3540-3549). JMLR. org.

157. Vezhnevets, A., Mnih, V., Osindero, S., Graves, A., Vinyals, O., and Agapiou, J. (2016). Strategic attentive writer for learning macro-actions. In Advances in neural information processing systems (pp. 3486-3494).

158. Vinitsky, E., Kreidieh, A., Le Flem, L., Kheterpal, N., Jang, K., Wu, C., Wu, F., Liaw, R., Liang, E. and Bayen, A.M. (2018). Benchmarks for reinforcement learning in mixed-autonomy traffic. In Conference on Robot Learning (pp. 399-409).

159. Vinitsky, E., Parvate, K., Kreidieh, A., Wu, C., and Bayen, A. (2018). Lagrangian control through deep-rl: Applications to bottleneck decongestion. In 2018 21st International Conference on Intelligent Transportation Systems (ITSC) (pp. 759-765). IEEE.

160. Vinyals, O., Fortunato, M., and Jaitly, N. (2015). Pointer networks. In Advances in neural information processing systems (pp. 2692-2700).

161. Wan, C. H., and Hwang, M. C. (2018). Value-based deep reinforcement learning for adaptive isolated intersection signal control. IET Intelligent Transport Systems, 12(9), 1005-1010.

162. Wang, C., Zhang, X., Cong, L., Li, J., and Zhang, J. (2019g). Research on intelligent collision avoidance decision-making of unmanned ship in unknown environments. Evolving Systems, 10(4), 649-658.

163. Wang, G., Hu, J., Li, Z., and Li, L. (2019a). Cooperative Lane Changing via Deep Reinforcement Learning. arXiv preprint arXiv:1906.08662.

164. Wang, P., Chan, C. Y., and Li, H. (2019b). Automated Driving Maneuvers under Interactive Environment based on Deep Reinforcement Learning. Accepted at the 98th Annual Meeting of the Transportation Research Board (TRB).

165. Wang, J., Zhang, Q., Zhao, D., and Chen, Y. (2019c). Lane Change Decision-making through Deep Reinforcement Learning with Rule-based Constraints. In 2019 International Joint Conference on Neural Networks (IJCNN) (pp. 1-6). IEEE.

166. Wang, P., and Chan, C. Y. (2017). Formulation of deep reinforcement learning architecture toward autonomous driving for on-ramp merge. In 2017 IEEE 20th International Conference on Intelligent Transportation Systems (ITSC) (pp. 1-6). IEEE.

167. Wang, P., and Chan, C. Y. (2018). Autonomous ramp merge maneuver based on reinforcement learning with continuous action space. arXiv preprint arXiv:1803.09203.

168. Wang, P., Chan, C. Y., and de La Fortelle, A. (2018a). A reinforcement learning based approach for automated lane change maneuvers. In 2018 IEEE Intelligent Vehicles Symposium (IV) (pp. 1379-1384). IEEE.

169. Wang, S., Jia, D., and Weng, X. (2018b). Deep reinforcement learning for autonomous driving. arXiv preprint arXiv:1811.11329.

170. Wang, P., Li, Y., Shekhar, S., and Northrop, W. F. (2019d). Actor-Critic based Deep Reinforcement Learning Framework for Energy Management of Extended Range Electric Delivery Vehicles. In 2019 IEEE/ASME International Conference on Advanced Intelligent Mechatronics (AIM) (pp. 1379-1384). IEEE.

171. Wang, P., Li, Y., Shekhar, S., and Northrop, W. F. (2019e). A deep reinforcement learning framework for energy management of extended range electric delivery vehicles. In 2019 IEEE Intelligent Vehicles Symposium (IV) (pp. 1837-1842). IEEE.

172. Wang, P., Liu, D., Chen, J., Li, H., and Chan, C. Y. (2020). Human-like Decision Making for Autonomous Driving via Adversarial Inverse Reinforcement Learning. arXiv, arXiv-1911.

173. Wang, R., Zhou, M., Li, Y., Zhang, Q., and Dong, H. (2019f). A Timetable Rescheduling Approach for Railway based on Monte Carlo Tree Search. In 2019 IEEE Intelligent Transportation Systems Conference (ITSC) (pp. 3738-3743). IEEE.

174. Wang, Z., Schaul, T., Hessel, M., Van Hasselt, H., Lanctot, M., and De Freitas, N. (2016). Dueling network architectures for deep reinforcement learning. arXiv preprint arXiv:1511.06581.

175. Wen, J., Zhao, J., and Jaillet, P. (2017). Rebalancing shared mobility-on-demand systems: A reinforcement learning approach. In 2017 IEEE 20th International Conference on Intelligent Transportation Systems (ITSC) (pp. 220-225). IEEE.

176. Williams, R. J. (1992). Simple statistical gradient-following algorithms for connectionist reinforcement learning. Machine learning, 8(3-4), 229-256.

177. Williams, R. J., and Peng, J. (1991). Function optimization using connectionist reinforcement learning algorithms. Connection Science, 3(3), 241-268.

178. Wolf, P., Kurzer, K., Wingert, T., Kuhnt, F., and Zollner, J. M. (2018). Adaptive behavior generation for autonomous driving using deep reinforcement learning with compact semantic states. In 2018 IEEE Intelligent Vehicles Symposium (IV) (pp. 993-1000). IEEE.

179. Wolf, P., Kurzer, K., Wingert, T., Kuhnt, F., and Zollner, J. M. (2018). Adaptive behavior generation for autonomous driving using deep reinforcement learning with compact semantic states. In 2018 IEEE Intelligent Vehicles Symposium (IV) (pp. 993-1000). IEEE.

180. Woo, J., Yu, C., and Kim, N. (2019). Deep reinforcement learning-based controller for path following of an unmanned surface vehicle. Ocean Engineering, 183, 155-166.

181. Wu, C., Parvate, K., Kheterpal, N., Dickstein, L., Mehta, A., Vinitsky, E., and Bayen, A. M. (2017). Framework for control and deep reinforcement learning in traffic. In 2017 IEEE 20th International Conference on Intelligent Transportation Systems (ITSC) (pp. 1-8). IEEE.

182. Wu, J., He, H., Peng, J., Li, Y., and Li, Z. (2018a). Continuous reinforcement learning of energy management with deep Q network for a power split hybrid electric bus. Applied energy, 222, 799-811.

183. Wu, Y., Tan, H., and Ran, B. (2018b). Differential Variable Speed Limits Control for Freeway Recurrent Bottlenecks via Deep Reinforcement learning. arXiv preprint arXiv:1810.10952.

184. Wu, Y., Tan, H., Peng, J., and Ran, B. (2019a). A Deep Reinforcement Learning Based Car Following Model for Electric Vehicle. 智能城市应用, 2(5).

185. Wu, Y., Tan, H., Peng, J., Zhang, H., and He, H. (2019b). Deep reinforcement learning of energy management with continuous control strategy and traffic information for a series-parallel plug-in hybrid electric bus. Applied energy, 247, 454-466.

186. Wulfmeier, M., Ondruska, P., and Posner, I. (2015). Maximum entropy deep inverse reinforcement learning. arXiv preprint arXiv:1507.04888.

187. Xu, Z., Tang, C., and Tomizuka, M. (2018). Zero-shot deep reinforcement learning driving policy transfer for autonomous vehicles based on robust control. In 2018 21st International Conference on Intelligent Transportation Systems (ITSC) (pp. 2865-2871). IEEE.

188. Yang, G., Zhang, F., Gong, C., and Zhang, S. (2019). Application of a Deep Deterministic Policy Gradient Algorithm for Energy-Aimed Timetable Rescheduling Problem. Energies, 12(18), 3461.

189. Ye, F., Cheng, X., Wang, P., and Chan, C. Y. (2020). Automated Lane Change Strategy using Proximal Policy Optimization-based Deep Reinforcement Learning. arXiv preprint arXiv:2002.02667.

190. Ye, Y., Zhang, X., and Sun, J. (2019). Automated vehicle's behavior decision making using deep reinforcement learning and high-fidelity simulation environment. Transportation Research Part C: Emerging Technologies, 107, 155-170.

191. Yi, H. (2018). Deep deterministic policy gradient for autonomous vehicle driving. In Proceedings on the International Conference on Artificial Intelligence (ICAI) (pp. 191-194). The Steering Committee of The World Congress in Computer Science, Computer Engineering and Applied Computing (WorldComp).
192. You, C., Lu, J., Filev, D., and Tsiotras, P. (2019). Advanced planning for autonomous vehicles using reinforcement learning and deep inverse reinforcement learning. Robotics and Autonomous Systems, 114, 1-18.
193. Yu, L., Shao, X., Wei, Y., and Zhou, K. (2018). Intelligent land-vehicle model transfer trajectory planning method based on deep reinforcement learning. Sensors, 18(9), 2905.
194. Zaheer, M., Kottur, S., Ravanbakhsh, S., Poczos, B., Salakhutdinov, R. R., and Smola, A. J. (2017). Deep sets. In Advances in neural information processing systems (pp. 3391-3401).
195. Zhang, Q. S., and Zhu, S. C. (2018). Visual interpretability for deep learning: a survey. Frontiers of Information Technology and Electronic Engineering, 19(1), 27-39.
196. Zhang, K., Li, M., Zhang, Z., Lin, X., and He, F. (2020). Multi-Vehicle Routing Problems with Soft Time Windows: A Multi-Agent Reinforcement Learning Approach. arXiv preprint arXiv:2002.05513.
197. Zhang, S., Peng, H., Nageshrao, S., and Tseng, E. (2019). Discretionary Lane Change Decision Making using Reinforcement Learning with Model-Based Exploration. In 2019 18th IEEE International Conference On Machine Learning And Applications (ICMLA) (pp. 844-850). IEEE.
198. Zhang, X., Wang, C., Liu, Y., and Chen, X. (2019). Decision-making for the autonomous navigation of maritime autonomous surface ships based on scene division and deep reinforcement learning. Sensors, 19(18), 4055.
199. Zhang, Z., Yang, J., and Zha, H. (2019). Integrating independent and centralized multi-agent reinforcement learning for traffic signal network optimization. arXiv preprint arXiv:1909.10651.
200. Zhao, L., and Roh, M. I. (2019). COLREGs-compliant multiship collision avoidance based on deep reinforcement learning. Ocean Engineering, 191, 106436.
201. Zhao, L., Roh, M. I., and Lee, S. J. (2019). Control method for path following and collision avoidance of autonomous ship based on deep reinforcement learning. Journal of Marine Science and Technology, 27(4), 293-310.

202. Zhao, P., Wang, Y., Chang, N., Zhu, Q., and Lin, X. (2018). A deep reinforcement learning framework for optimizing fuel economy of hybrid electric vehicles. In 2018 23rd Asia and South Pacific Design Automation Conference (ASP-DAC) (pp. 196-202). IEEE.

203. Zhou, M., Yu, Y., and Qu, X. (2019). Development of an efficient driving strategy for connected and automated vehicles at signalized intersections: A reinforcement learning approach. IEEE Transactions on Intelligent Transportation Systems.

204. Zhou, R., and Song, S. (2018). Optimal automatic train operation via deep reinforcement learning. In 2018 Tenth International Conference on Advanced Computational Intelligence (ICACI) (pp. 103-108). IEEE.

205. Zhou, R., Song, S., Xue, A., You, K., and Wu, H. (2020). Smart Train Operation Algorithms based on Expert Knowledge and Reinforcement Learning. arXiv preprint arXiv:2003.03327.

206. Zhu, F., and Ukkusuri, S. V. (2014). Accounting for dynamic speed limit control in a stochastic traffic environment: A reinforcement learning approach. Transportation research part C: emerging technologies, 41, 30-47.

207. Zhu, L., He, Y., Yu, F. R., Ning, B., Tang, T., and Zhao, N. (2017). Communication-based train control system performance optimization using deep reinforcement learning. IEEE Transactions on Vehicular Technology, 66(12), 10705-10717.

208. Zhu, M., Wang, X., and Wang, Y. (2018). Human-like autonomous car-following model with deep reinforcement learning. Transportation research part C: emerging technologies, 97, 348-368.

209. Zhu, M., Wang, Y., Pu, Z., Hu, J., Wang, X., and Ke, R. (2019). Safe, Efficient, and Comfortable Velocity Control based on Reinforcement Learning for Autonomous Driving. arXiv preprint arXiv:1902.00089.
60